\documentclass[10pt,twocolumn,letterpaper]{article}

\usepackage{iccv}              
\usepackage[accsupp]{axessibility}
\usepackage{xcolor}
\usepackage{booktabs}
\usepackage{caption}
\usepackage{floatrow}
\usepackage{mathtools}
\allowdisplaybreaks

\definecolor{iccvblue}{rgb}{0.21,0.49,0.74}
\definecolor{text_red}{RGB}{238,124,146}
\definecolor{image_blue}{RGB}{114,161,219}
\usepackage[pagebackref,breaklinks,colorlinks,allcolors=iccvblue]{hyperref}
\usepackage{appendix}
\AtBeginEnvironment{appendices}{\crefalias{section}{appendix}}

\def\paperID{5642} 
\def\confName{ICCV}
\def\confYear{2025}

\title{TF-TI2I: Training-Free Text-and-Image-to-Image Generation via Multi-Modal Implicit-Context Learning in Text-to-Image Models}

\author{Teng-Fang Hsiao, Bo-Kai Ruan, Yi-Lun Wu, Tzu-Ling Lin, Hong-Han Shuai, \\
National Yang Ming Chiao Tung University\\
{\tt\small \{tfhsiao.ee13, bkruan.ee11, yilun.ee08, tzulinglin.11, hhshuai\}@nycu.edu.tw}
}
\begin{document}

\twocolumn[{%
\renewcommand\twocolumn[1][]{#1}%
\maketitle
\begin{center}
    \centering
    \captionsetup{type=figure}
    \includegraphics[width=\linewidth]{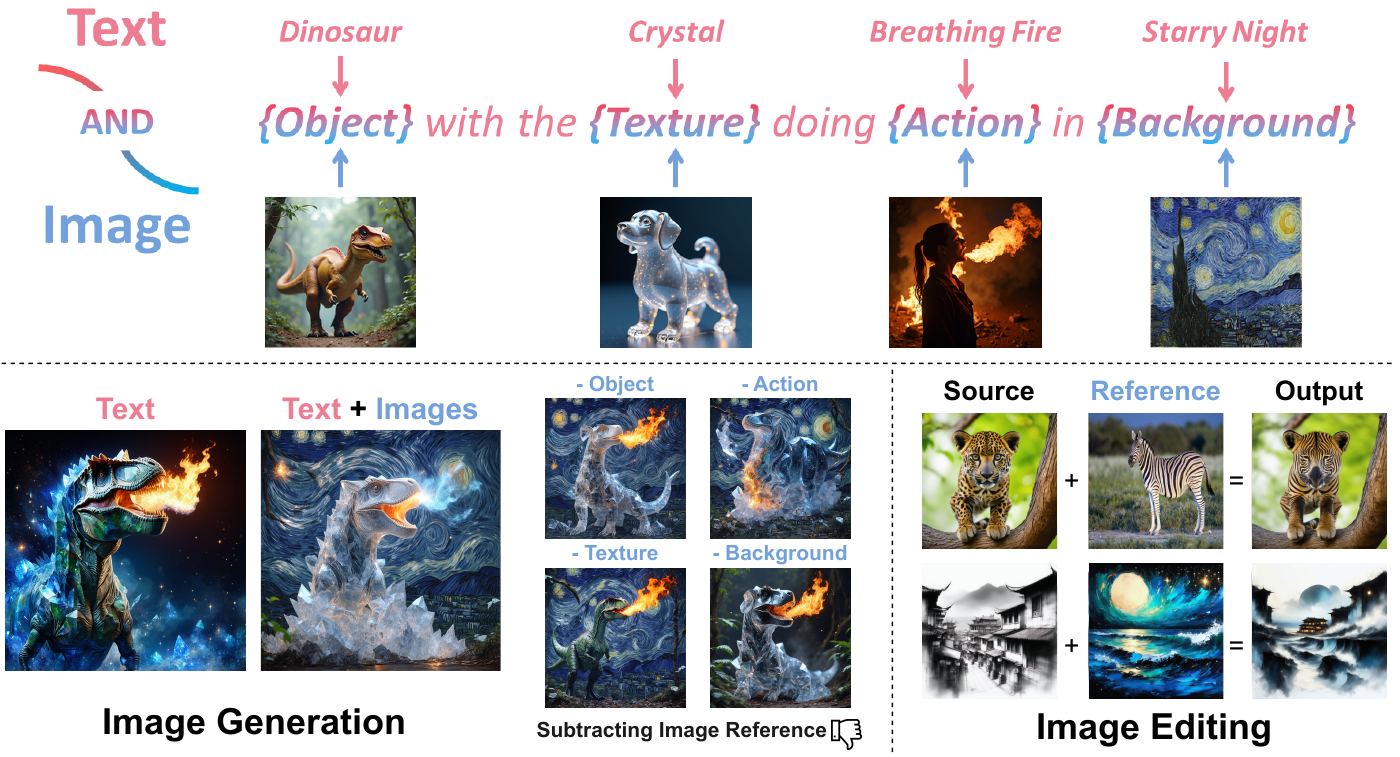}
    \caption{Illustration of our proposed TF-TI2I, designed to leverage multiple image references for generation and editing without additional training. The importance of image references is illustrated in the middle part, where excluding any reference significantly alters the output.}  
    \label{fig:teaser}
    \end{center}%
    }]
\maketitle
\begin{abstract}
Text-and-Image-To-Image (TI2I), an extension of Text-To-Image (T2I), integrates image inputs with textual instructions to enhance image generation. Existing methods often partially utilize image inputs, focusing on specific elements like objects or styles, or they experience a decline in generation quality with complex, multi-image instructions. To overcome these challenges, we introduce Training-Free Text-and-Image-to-Image (TF-TI2I), which adapts cutting-edge T2I models such as SD3 without the need for additional training. Our method capitalizes on the MM-DiT architecture, in which we point out that textual tokens can implicitly learn visual information from vision tokens. We enhance this interaction by extracting a condensed visual representation from reference images, facilitating selective information sharing through Reference Contextual Masking—this technique confines the usage of contextual tokens to instruction-relevant visual information. Additionally, our Winner-Takes-All module mitigates distribution shifts by prioritizing the most pertinent references for each vision token. Addressing the gap in TI2I evaluation, we also introduce the FG-TI2I Bench, a comprehensive benchmark tailored for TI2I and compatible with existing T2I methods. Our approach shows robust performance across various benchmarks, confirming its effectiveness in handling complex image-generation tasks. Our project page: \url{https://bluedyee.github.io/TF-TI2I_page/}
\end{abstract}    
\section{Introduction}
\label{sec:intro}

\noindent\textit{``Don’t tell me the moon is shining; show me the glint of light on broken glass.''}

\hfill— Anton Chekhov (1860-1904)

Recent advances in text-to-image (T2I) generation empower users to produce compelling visual artwork using only textual prompts~\cite{sd,sdxl,sana,flux,janusflow}. However, text descriptions alone are insufficient for conveying the nuanced spatial relationships, specific object characteristics, and stylistic details necessary to fulfill users’ intentions. To address this, various approaches have integrated additional visual guidance into T2I pipelines, leveraging conditional-control methods~\cite{controlnet,uni_controlnet,smartcontrol}, inpainting techniques~\cite{brushnet,powerpaint,smartbrush,freecond}, style transfer~\cite{style_injection,tfgph,zstar,stylebooth}, and customized image synthesis~\cite{dreambooth,anydoor,textual_inversion,ip_adapter,ms_diffusion}.

Despite their effectiveness, current methods are typically constrained to a single functionality. For example, conditional-control, inpainting, and style-transfer approaches focus on one aspect of visual guidance. Similarly, customization methods excel at object-based references but struggle with other factors \,  e.g., texture or background. One solution is the use of multimodal large language models (MLLMs) in image generation~\cite{kosmosg,emu2}, which can handle both visual and textual inputs. However, MLLM-based generation still lags behind specialized T2I models in terms of output quality and resolution, largely due to the limited availability of text-and-image-to-image training data. For instance, KOSMOS-G~\cite{kosmosg}, utilizing BLIP2~\cite{blip2} and CLIP-Seg~\cite{blip2} to autonomously identify objects as image inputs, remains restricted to object-level conditioning without providing finer control over texture or background details.

To enhance performance without larger training data, we explore MM-DiT—a fully transformer-based multimodal architecture used in state-of-the-art T2I models such as SD3~\cite{sd3} and FLUX~\cite{flux}. We contend that MM-DiT’s multimodal attention demonstrates cross-modal understanding~\cite{emergent_correspondance,unsupervised_semantic_correspondence,generative_what}, which allows textual tokens to naturally integrate \textbf{implicit visual information} from image latent during inference. Unlike methods that require additional training to establish image-reference correlations, we propose \textbf{Training-Free Text-and-Image-to-Image (TF-TI2I)} by (1) harnessing the inherent ability of MM-DiT to infuse textual tokens with implicit visual context during generation, and (2) enabling these tokens to be shared across different inputs as contextual tokens. As our approach relies only on the model's intrinsic properties, we can extend T2I models with TI2I capabilities without fine-tuning.

However, challenges arise as the number of image references increases. Firstly, mutual interference between different references can lead to unintended visual blending. For instance, as shown in Fig.~\ref{fig:teaser}, when generating a background following reference 4 (Starry Night), elements from the backgrounds of references 1, 2, and 3 might inadvertently be incorporated due to insufficient differentiation among the references. To address the mutual interference among multiple references, we propose \textbf{References Contextual Masking (RCM)}. This technique limits the visual information learned from contextual tokens by focusing exclusively on vision tokens that are more related to the given instructions. This ensures that only the relevant visual information of a specified reference is utilized in the image generation process. By doing so, RCM effectively reduces the undesired blending of visual features from different references, maintaining clarity and fidelity to the original input conditions.

Secondly, as a training-free approach, TF-TI2I is also susceptible to distribution shifts that can degrade generation quality, a challenge noted in related techniques~\cite{style_injection,freeenhance,zstar}. To counteract this, we introduce the \textbf{Winner-Takes-All} (WTA) module. This module assigns each vision token exclusively to one reference at a time, ensuring that each token robustly represents the visual characteristics of its assigned references. Concurrently, the WTA module facilitates the incorporation of visual information from other references through attention between the visual tokens. This dual mechanism not only maintains each reference’s integrity within the output but also enriches the visual details, resulting in high-quality images that adhere to both intended references and textual instruction.

Finally, recognizing the need for a comprehensive evaluation framework for both TI2I models~\cite{emu2,kosmosg} and visually guided T2I models~\cite{omnigen,masactrl,style_align,controlnet,customcontrast}, we introduce the Fine-Grained TI2I Benchmark (FG-TI2I Bench) tailored for general image generation scenarios. Drawing inspiration from EditBench~\cite{editbench}, which assesses detailed object attributes like material, color, and shape, our benchmark structures prompt instructions around four key components and encompass 6 fundamental TI2I sub-tasks. These components are described using either text or images, making our framework compatible with visually guided T2I methods. Our approach achieves state-of-the-art performance on 12 out of 18 evaluation metrics across general-purpose TI2I tasks and remains highly competitive in task-specific benchmarks such as DreamBench~\cite{dreambooth} for customization and Wild-TI2I~\cite{pnp} for editing scenarios.

Our contributions are summarized as follows:
\begin{itemize}
    \item We discover the implicit-context learning capability of MM-DiT and design TF-TI2I, a training-free approach, to generate images from different visual references and text prompts. This further augments the existing T2I model with TI2I capability.
    \item We develop \textit{Reference Contextual Masking} and \textit{Winner Takes All} modules to mitigate multi-reference problems, leading to more visually satisfactory results.
    \item We introduce FG-TI2I Bench, a comprehensive benchmark designed to evaluate a variety of visually guided T2I application \eg customization and style transfer. Our approach achieves state-of-the-art results in 12 out of 18 evaluation metrics across FG-TI2I and also remains competitive in DreamBench~\cite{dreambooth} and Wild-TI2I~\cite{pnp}. 
\end{itemize}

\section{Related Work}
\label{sec:related_work}

\subsection{TI2I with single aspect of reference}  
To expand the creativity of pretrained T2I models with TI2I capability, finetuning-based approaches~\cite{dreambooth,ip_adapter,stylebooth,textual_inversion,vct,lambda_eclipse}, such as DreamBooth~\cite{dreambooth}, encode reference information into model weights through a finetuning process. While effective, these methods are often limited by computational costs and the need for a substantial number of training references. In contrast, training-free methods~\cite{tf_icon,tfgph,style_injection,zstar,customcontrast} provide efficiency but concentrate on a single aspect of reference images, such as style in StyleID~\cite{style_injection} and ZSTAR~\cite{zstar}, or objects in TF-ICON~\cite{tf_icon} and TF-GPH~\cite{tfgph}. Our approach, which integrates the \textit{Reference Context Control} and \textit{Winner-Takes-All} modules, efficiently combines various aspects of references—including objects, textures, actions, and backgrounds—enabling a more comprehensive and flexible generation capability.

\subsection{TI2I with multiple aspects of references}  
Models trained from scratch with both text and image inputs can better utilize diverse image references for TI2I tasks. These models can be broadly categorized into retrieval-based~\cite{retrieve_diffusion,re_imagen,knn_diffusion} and auto-regressive-based approaches~\cite{emu2,kosmosg,omnigen}. Retrieval-based methods, such as KNN-Diffusion~\cite{knn_diffusion} and RDM~\cite{retrieve_diffusion}, follow the concept of RAG~\cite{RAG} by retrieving relevant images from a database to provide visual prior for generation. However, their capability is dependent on the retrieved references, which limits the model’s ability to synthesize novel content beyond what is provided. In contrast, auto-regressive-based methods like Emu2~\cite{emu2} and Kosmos-G~\cite{kosmosg} face constraints in generation resolution and quality due to the limited availability of multimodal training pairs. Our proposed TF-TI2I leverages a pretrained T2I model, enhancing it with reference support for controllability, and achieves superior performance over existing methods without additional training costs.

\section{Preliminary}
\label{sec:preliminary}
Our work leverages the Multimodal Attention (MMA) blocks inherent in MM-DiT-based T2I models~\cite{sd3,flux}. Given a textual prompt $P$ and a reference image $I$, the prompt and image are first encoded into latent features, denoted as $\tau_{P}^{0} = \mathcal{E}_{P}(P) \in \mathbb{R}^{n_p \times d }$ and $\tau_{I}^0 = \mathcal{E}_{I}(I) \in \mathbb{R}^{n_{I} \times d}$, where $\mathcal{E}_{P}$ and $\mathcal{E}_{I}$ represent the text and image encoders, respectively. For the latent features within the MM-DiT backbone, we use the superscript $l \in {0,1,\dots, L}$ to indicate the layer, e.g.\@, $\tau_{I}^{l}$ and $\tau_{P}^{l}$. For simplicity, we omit the time step notation, as it remains unchanged in all equations, i.e.\@, $\tau_{I}^{l} = \tau_{I}^{l}(t)$. The multimodal attention mechanism, after disregarding the head dimension, is expressed as follows: 
\begin{align}
    \{Q_{I}, K_{I}, V_{I}\} &= \tau_{I}^{l} W_{I}^{\{Q,K,V\}}, \\
    \{Q_{P}, K_{P}, V_{P}\} &= \tau_{P}^{l} W_{P}^{\{Q,K,V\}}, \\
    \text{A}(\tau_{I}^{l},\tau_{P}^{l},M) &= \text{softmax}\big((QK^T+M)/\sqrt{d}\big)V, 
    \label{eq:MMA}
\end{align}
where $Q = [Q_{I}; Q_{P}]$, $K = [K_{I}; K_{P}]$, and $V = [V_{I}; V_{P}]$, with $[\cdot; \cdot]$ denoting concatenation along the first dimension. Here, $W$ represents the projection matrix, and $M$ denotes the attention mask. A key distinction of MM-DiT from previous self- and cross-attention~\cite{sd,sdxl} is that it updates not only the visual tokens $\tau_{I}^{l}$ but also refines the textual tokens $\tau_{P}^{l}$. This dual-update process is formulated as:
\begin{equation}
    [\tau_{I}^{l+1}; \tau_{P}^{l+1}] = [\tau_{I}^{l}; \tau_{P}^{l}] + \text{A}(\tau_{I}^{l}, \tau_{P}^{l}, M).
\end{equation}
Throughout this paper, we use SD3.5-large~\cite{sd3} as our default backbone, which consists of 38 MM-DiT layers, with $n_{I} = 4096$ and $n_{P} = 333$. Additionally, to facilitate discussions on the diffusion process in subsequent sections, we incorporate the noise-adding process, defined as
\begin{equation}
    \tau_{I}^{0}(t) = \tau_{I}^{0} + \epsilon(t),
\end{equation}
where $\epsilon$ represents either standard Gaussian noise or inversion-based noise predictors~\cite{ddim,dpm_sovler_plus,rf_inversion,lightningfast_inversion}.


\begin{figure*}
    \centering
    \includegraphics[width=\linewidth]{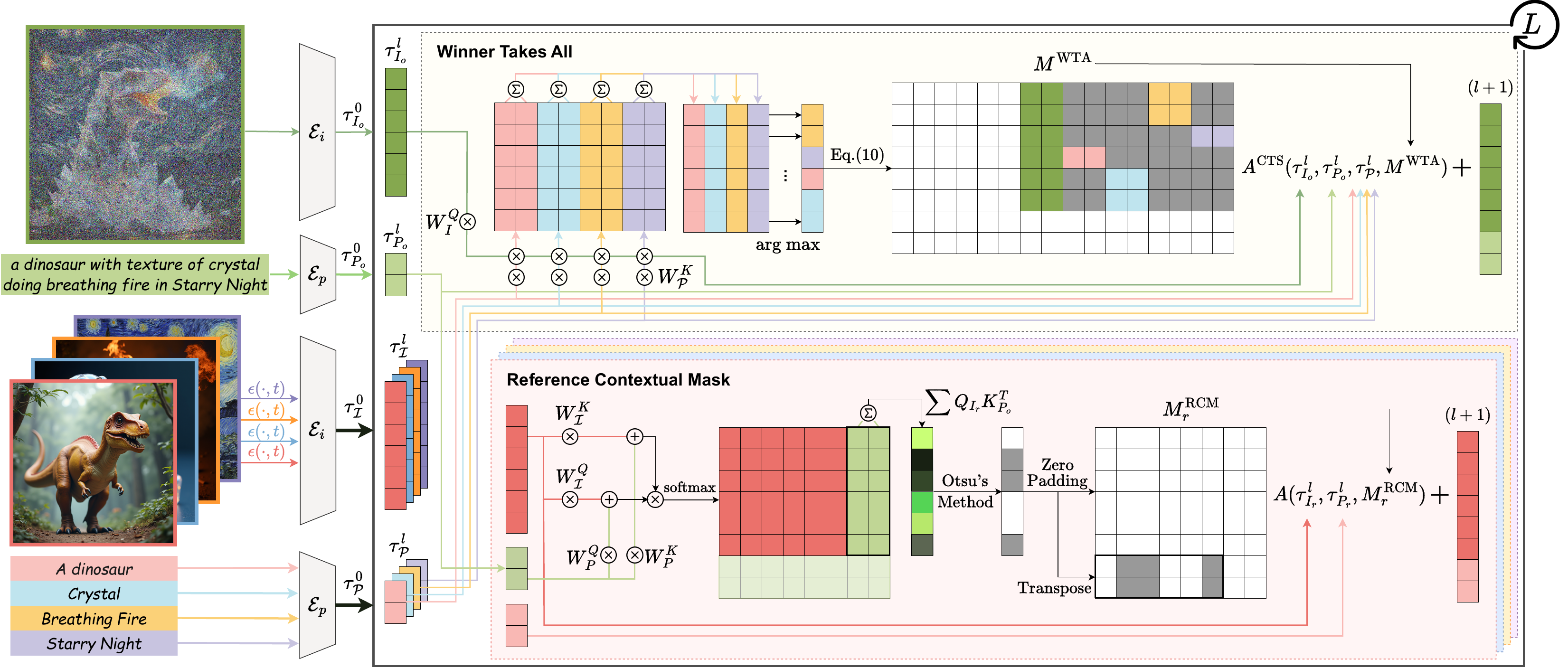}
    \caption{Illustration of the TF-TI2I pipeline. TF-TI2I leverages contextual visual information learned from textual tokens. Through sharing the contextual token by concatenating $\tau_{\mathcal{P}}^{l}$ to the upper block (\cref{sec:CTS}), we achieve prompt-following while maintaining reference-aligned results. Additionally, we incorporate Reference Contextual Masking (\cref{sec:RCM}) to mitigate mutual interference between references and employ the Winner-Takes-All module (\cref{sec:WTA}) to minimize distribution shifts in multi-reference scenarios.}
    \label{fig:pipline}
\end{figure*}

\section{Method}
\label{sec:method}
In this section, we first examine the implicit-learning capability of textual tokens. Next, we demonstrate that sharing these textual tokens enables the manipulation of visual content through textual representations. Finally, we propose References Contextual Masking and Winner Takes All module to extract and assign instruction-relevant information from excessively rich visual content. The overall architecture can be found in \cref{fig:pipline}.

\begin{figure}[htbp]
\footnotesize
    \centering
    \begin{subfigure}[c]{0.22\linewidth}{\includegraphics[width=\linewidth]{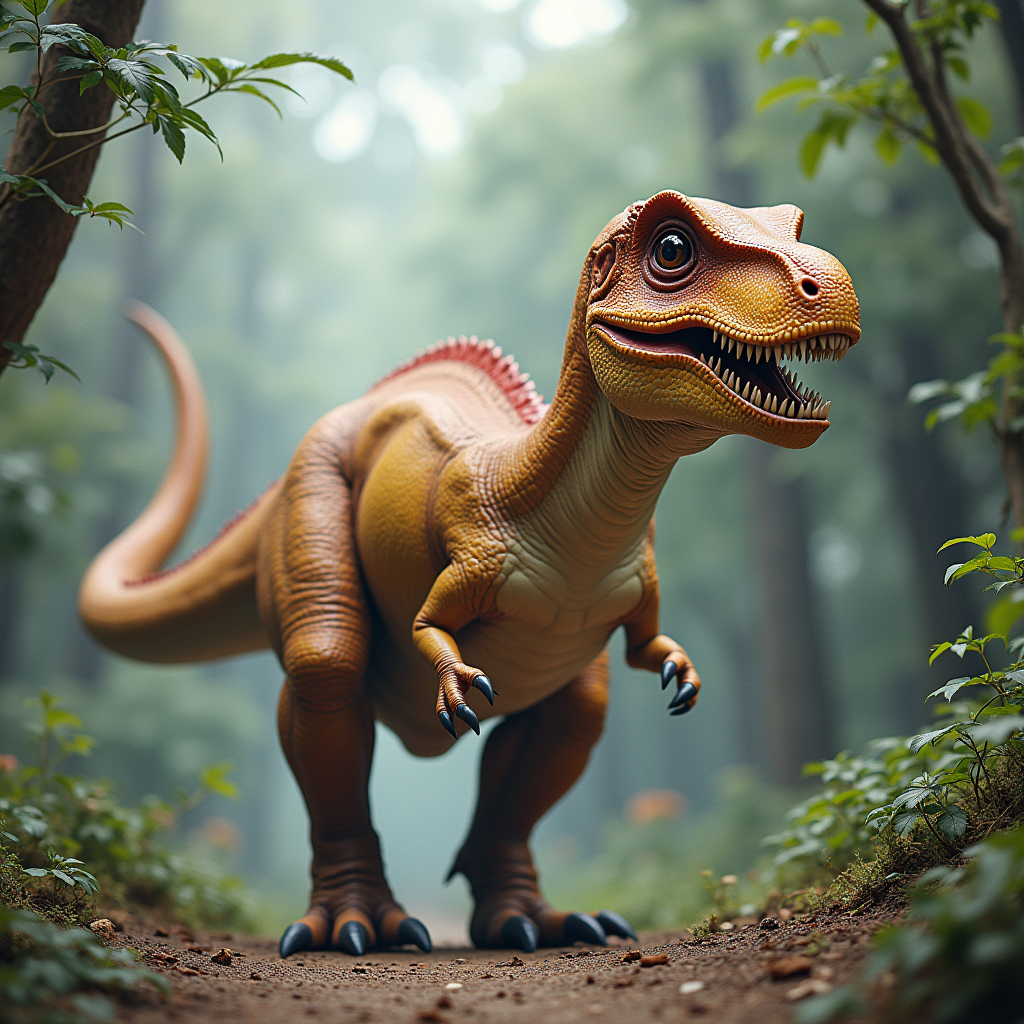}}
    \caption{$I_1$}
    \label{fig:tsne:a}
    \end{subfigure}
    \begin{subfigure}[c]{0.06\linewidth}
    \hfill
    \rotatebox{90}{ \makebox{$t=0.8T$}}
    \hfill
    \caption*{}
    \end{subfigure}
    \begin{subfigure}[c]{0.22\linewidth}{\includegraphics[width=\linewidth]{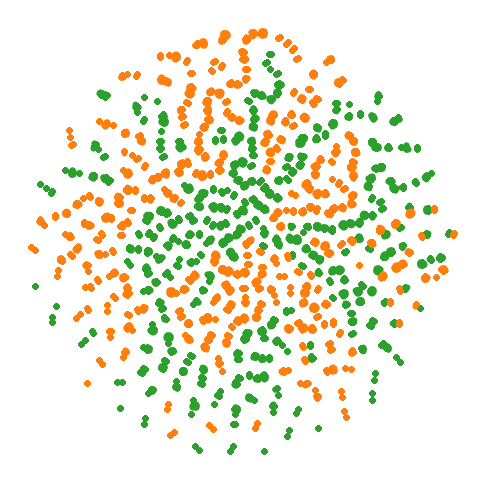}}
    \caption*{}
    \label{fig:tsne:f}
    \end{subfigure}
    \begin{subfigure}[c]{0.22\linewidth}{\includegraphics[width=\linewidth]{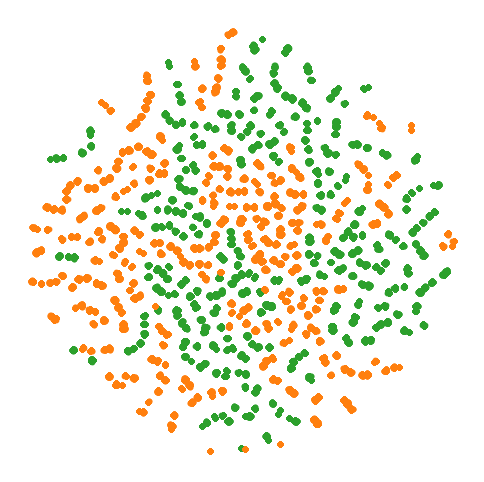}}
    \caption*{}
    \label{fig:tsne:g}
    \end{subfigure}
    \begin{subfigure}[c]{0.22\linewidth}{\includegraphics[width=\linewidth]{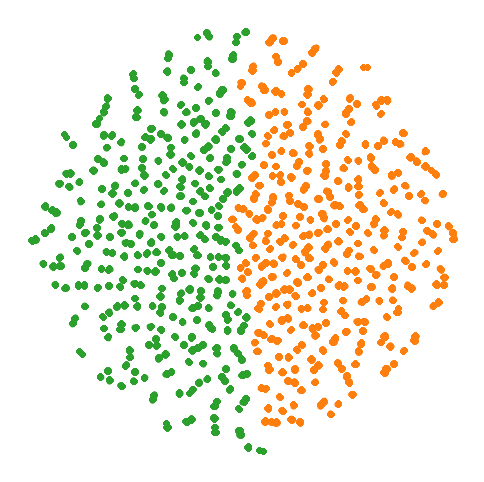}}
    \caption*{}
    \label{fig:tsne:h}
    \end{subfigure}
    
    \begin{subfigure}[c]{0.22\linewidth}{\includegraphics[width=\linewidth]{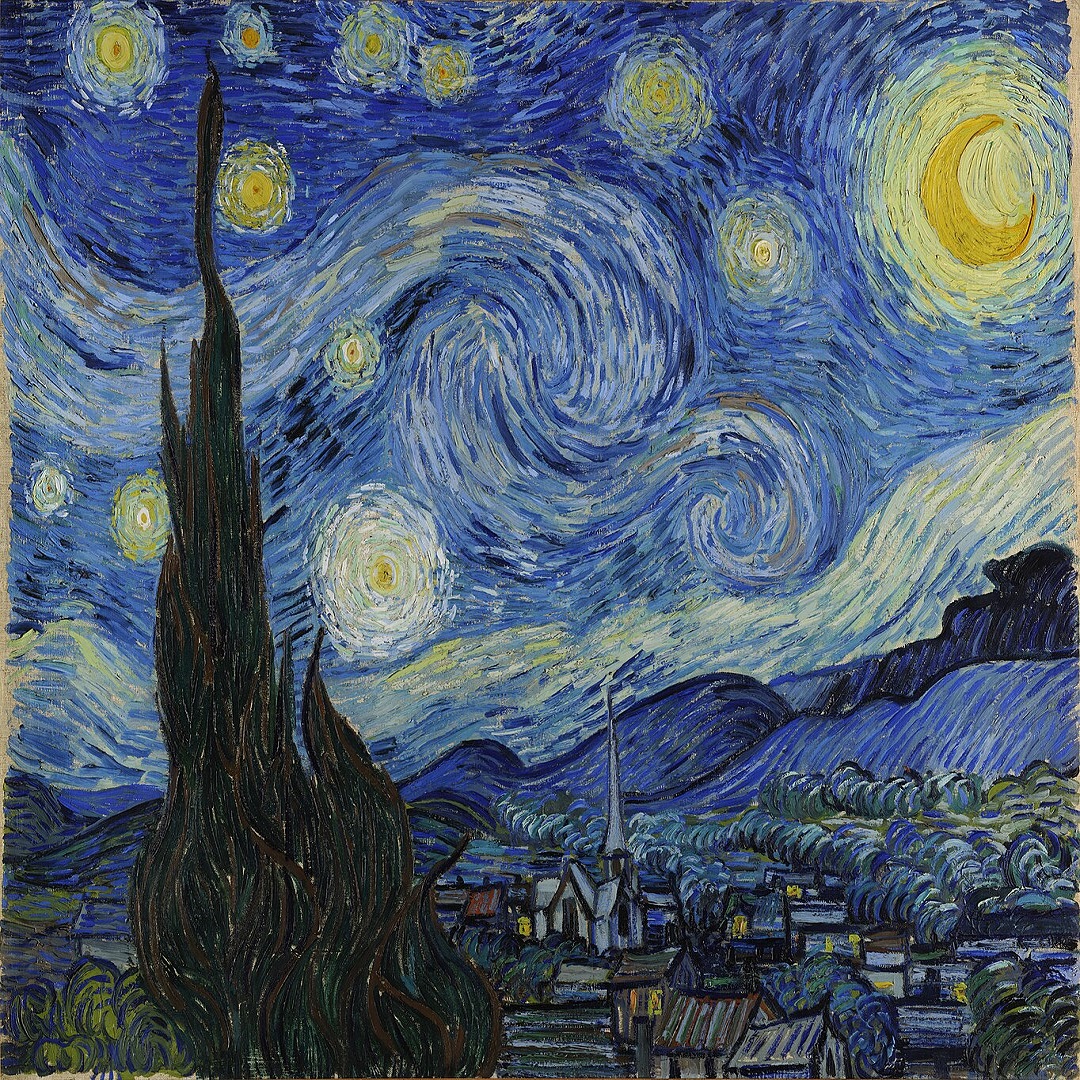}}
    \caption{$I_2$}
    \label{fig:tsne:b}
    \end{subfigure}
    \begin{subfigure}[c]{0.06\linewidth}
    \hfill
    \rotatebox{90}{ \makebox{$t=0.5T$}}
    \hfill
    \caption*{}
    \end{subfigure}
    \begin{subfigure}[c]{0.22\linewidth}{\includegraphics[width=\linewidth]{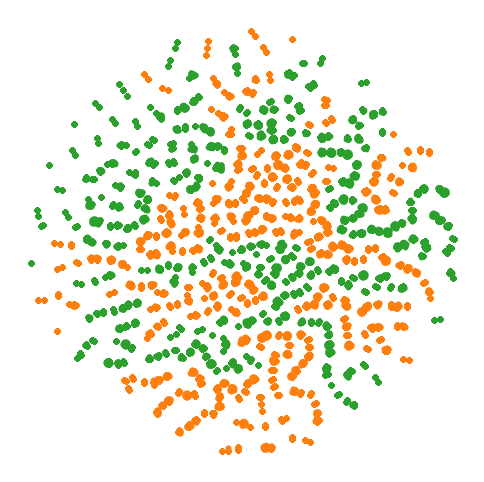}}
    \caption*{$l=1$}
    \end{subfigure}
    \begin{subfigure}[c]{0.22\linewidth}{\includegraphics[width=\linewidth]{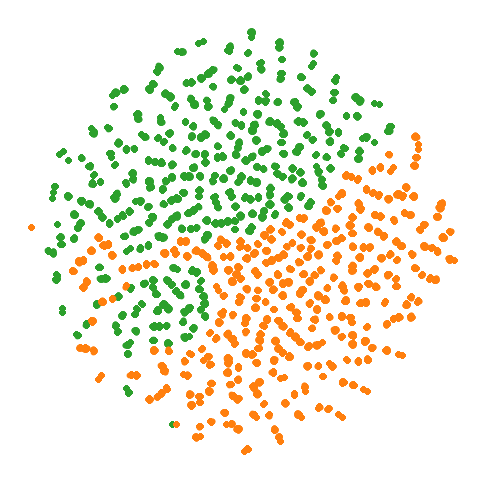}}
    \caption*{$l=18$}
    \label{fig:tsne:c}
    \end{subfigure}
    \begin{subfigure}[c]{0.22\linewidth}{\includegraphics[width=\linewidth]{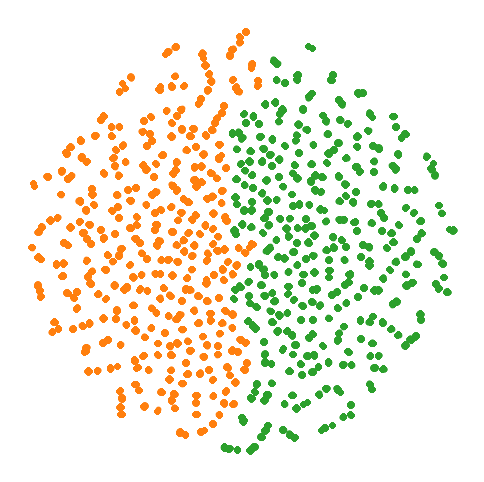}}
    \caption*{$l=37$}
    \label{fig:tsne:d}
    \end{subfigure}
    
\caption{The t-SNE visualization of $\{\textcolor[HTML]{2CA02C}{\tau^{l}_{P_1}} \mid P_1 \in \mathbb{P}\}$ and $\{\textcolor[HTML]{FF7F0E}{\tau^{l}_{P_2}} \mid P_2 \in \mathbb{P}\}$ at different timesteps and layers indices. The clusters form in deeper layers and later timesteps.}
\label{fig:tsne}

\end{figure}


\subsection{Implicit-Context Learning from MM-DiT}
\label{sec:incontext_learning}
The key distinction between MM-DiT and conventional T2I models~\cite{sd,sdxl,sdxl_lightning,pixart_alpha,pixart_delta,pixart_sigma}, lies in how textual information is treated. While traditional models use textual instruction as a fixed condition, MM-DiT updates the textual conditions $\tau_{P}^{l}$ at each layer. This characteristic motivates us to investigate the problem--\textbf{Is visual context learned implicitly by the textual tokens $\tau_{P}^{l}$ in the MM-DiT?} We use the term ``implicit'' to emphasize that textual tokens gradually aggregate visual context from visual tokens through multi-modal attention. Consequently, these textual tokens can implicitly represent the visual reference.

To empirically test whether textual tokens from different prompts encode similar features when conditioned on the same image, we sample 50 prompts structured as \textit{``\{Object\} with  the texture of {Texture} doing  \{Action\}  in the background of \{Background\}.''} We denote these prompts as $\mathbb{P}$ and consider two different images, $I_1$ (\cref{fig:tsne:a}) and $I_2$ (\cref{fig:tsne:b}). We then forward these prompt tokens, $\tau_{P_1}^{0} = \mathcal{E}_{P}(P_1)$ and $\tau_{P_2}^{0} = \mathcal{E}_{P}(P_2)$ for all $P_1, P_2 \in \mathbb{P}$, along with the vision tokens $\tau_{I_1}^{l}$ and $\tau_{I_2}^{l}$. Subsequently, we extract $\tau^{l}_{P_1}$ and $\tau^{l}_{P_2}$ from the T2I model to perform clustering visualization for different layer index $l$. Our hypothesis is that \textbf{if textual tokens learn from vision tokens during the forward process, the features distributions, $\{\tau^{l}_{P_1} \mid P_1 \in \mathbb{P}\}$ and $\{\tau^{l}_{P_2} \mid P_2 \in \mathbb{P}\}$, should form distinct clusters due to the different input visual tokens}.  As illustrated in the t-SNE visualization in ~\cref{fig:tsne}, 
we observe that $\tau_{P}^{l}$ exhibits stronger clustering effects in deeper layers and at later stages of generation, where visual features become more distinct from random noise. These findings highlight the implicit context-learning capability of the textual tokens $\tau_{P}$.

\begin{figure}[htbp]
\footnotesize

    \begin{subfigure}[b]
    {0.32\linewidth}{\includegraphics[width=\linewidth]{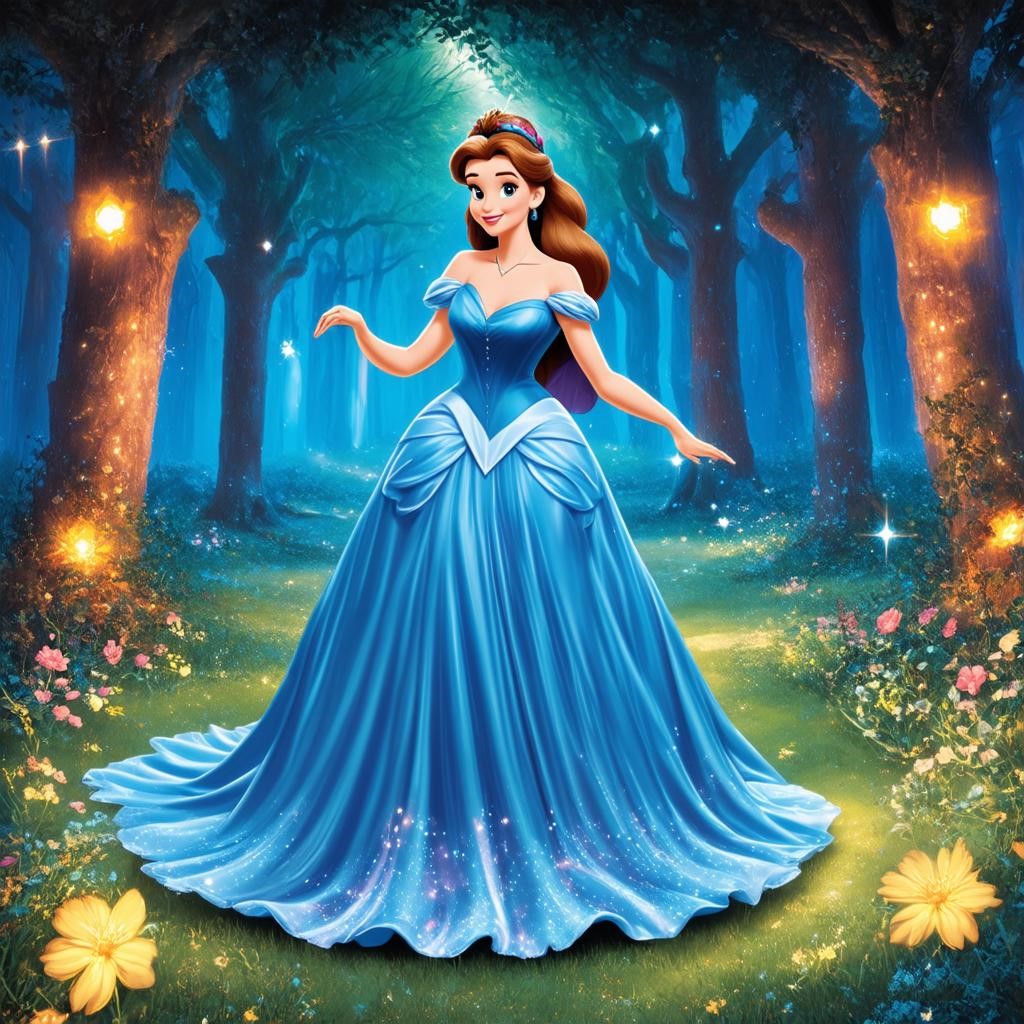}}
    \caption{$I_1$ from seed 1}
    \label{fig:replace:a}
    \end{subfigure}
    \begin{subfigure}[b]{0.32\linewidth}{\includegraphics[width=\linewidth]{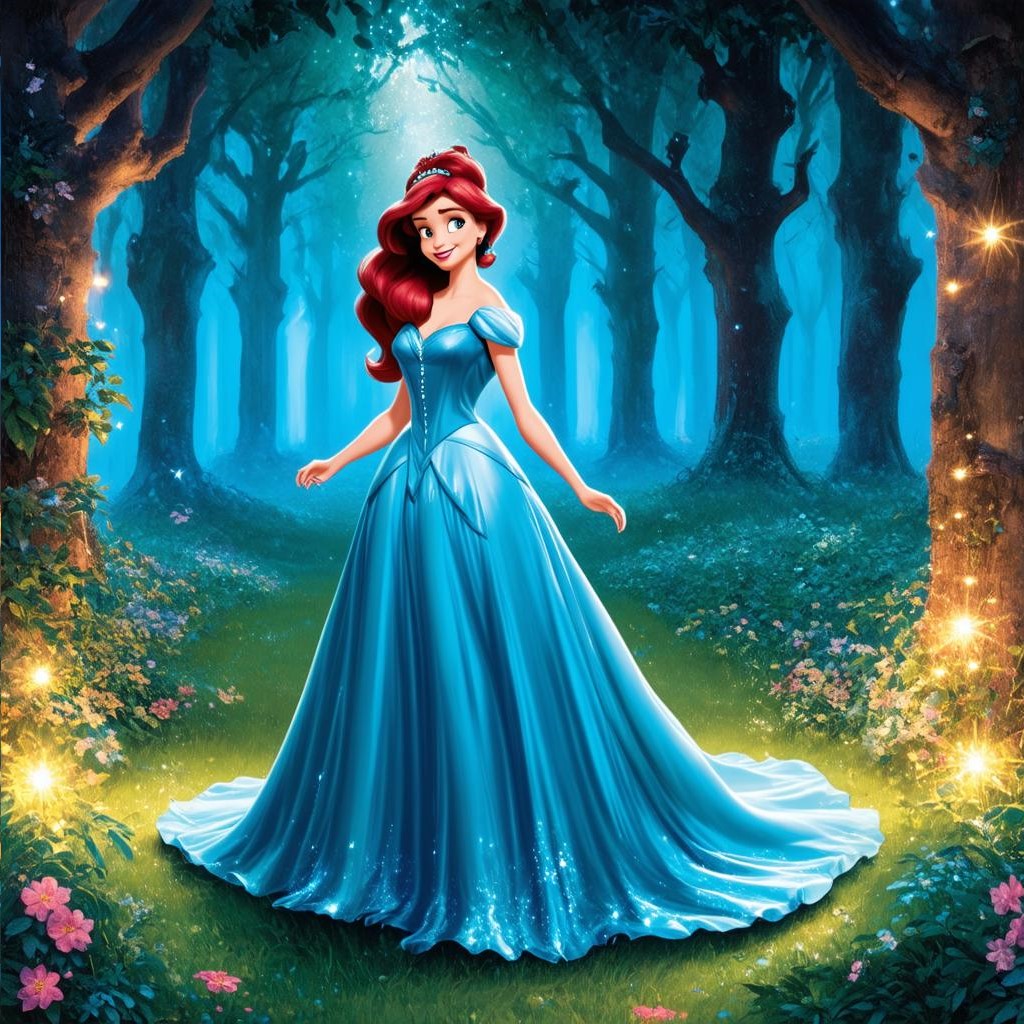}}
    
    \caption{$I_2$ from seed 2}
    \label{fig:replace:b}
    \end{subfigure}
    \begin{subfigure}[b]{0.32\linewidth}{\includegraphics[width=\linewidth]{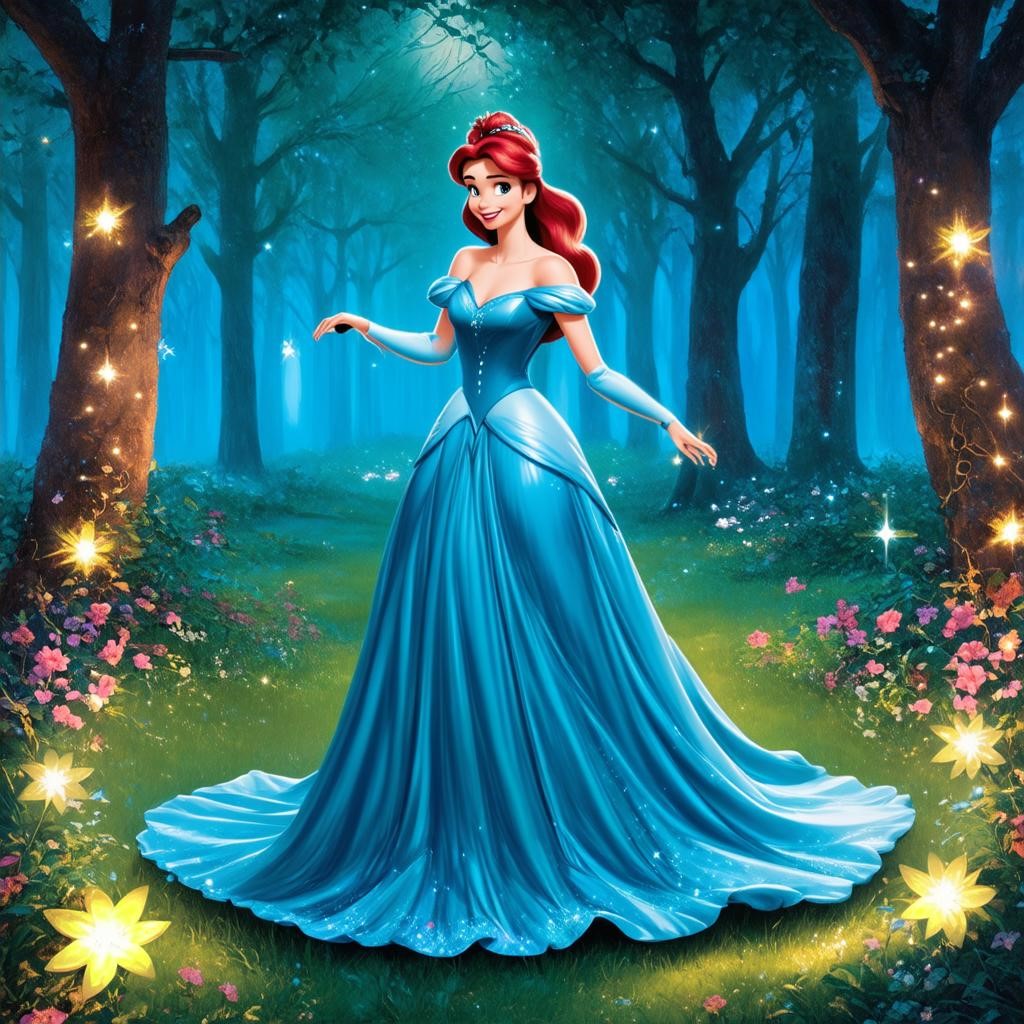}}

    \caption{Replacing $\tau_{P_1}$}
    \label{fig:replace:c}
    \end{subfigure}

\caption{The illustration of replacing contextual tokens between different images. As shown in the~\cref{fig:replace:c}, we can successfully transfer the visual information of~\cref{fig:replace:b} to ~\cref{fig:replace:a}.}
\label{fig:replace}

\end{figure}

In the following, we refer to these textual tokens with implicit-context visual information as \textit{contextual tokens} to distinguish them from the so-called textual tokens, which only contain textual information as described in prior work~\cite{sd,sdxl,sd3}. Our next question is: \textbf{Can the contextual tokens utilized by different set of vision tokens?} To test this possibility, we create a toy example where we use a single prompt, \textit{``a princess in the dress''}, with two different seeds. Due to the difference in the initial noise of the vision tokens, the generated images are different in both details and layout, as shown in~\cref{fig:replace:a} and \cref{fig:replace:b}. We then replace their contextual token by replacing $\tau^{l}_{P_1}$ with $\tau^{l}_{P_2}$ in each layer for all timesteps. The result is shown in~\cref{fig:replace:c}, where the image structure still resembles $I_1$, but features like dress color shift toward $I_2$. This example supports our hypothesis that visual information in contextual tokens can be utilized by other set of vision tokens.

\subsection{Training-Free Text-and-Image-to-Image}
Building upon the previous findings that contextual tokens not only encapsulate visual information but can also be leveraged by other vision tokens, we propose TF-TI2I, a Training-Free Text-and-Image-to-Image pipeline, as shown in \cref{fig:pipline}. By modifying pre-trained T2I models~\cite{sd3,flux}, we enable the utilization of multiple images as reference inputs while maintaining high-quality generation results.

\subsection{TF-TI2I Overview}
The objective of TF-TI2I is to generate an image $I_o$ based on the prompt instruction $P_o$ and given visual references $\mathcal{I} \coloneqq \{I_1, I_2, \dots, I_R\}$. At each timestep $t$, we start with an initial vision token $\tau^{0}_{I_o}$, which can be either a random noise for image generation or a noisy image for editing. Along with this, we incorporate the textual token $\tau_{P_o}^{0}$ derived from the input prompt $P_o$. We then introduce reference vision tokens $\{\tau^{0}_{I_1}, \tau^{0}_{I_2}, \dots, \tau^{0}_{I_R}\}$, where  $\tau^{0}_{I_r}(t) = \tau^{0}_{I_r} + \epsilon( t)$, with $\tau^{0}_{I_r} = \mathcal{E}_{I}(I_r)$ and $I_r \in \mathcal{I}$. Additionally, TF-TI2I requires reference prompts $\mathcal{P} \coloneqq \{P_1, P_2, \dots, P_R\}$, which are used to initialize the textual tokens $\{\tau_{P_1}^{0}, \tau_{P_2}^{0}, \dots, \tau_{P_R}^{0}\}$, where $\tau^{0}_{P_r} = \mathcal{E}_{P}(P_r)$ and $P_r \in \mathcal{P}$.


\subsubsection{Contextual Tokens Sharing (CTS)}
\label{sec:CTS}
Extending from the findings that contextual tokens can be utilized by other vision tokens, we design a contextual token-sharing module to replace the original MMA layer and leverage the contextual information from image references. The idea of CTS is akin to the share-attention module~\cite{masactrl,style_align,style_injection,tfgph}, which concatenates the vision tokens from different images together. By contrast, we concatenate the key and value of $\tau_{I_o}^{l}$ with contextual tokens $\tau_{P_r}^{l}$ to obtain the new input prompt tokens $\tau_{\mathcal{P}}^{l} := [\tau_{P_o}^{l}; \tau_{P_1}^{l}; \tau_{P_2}^{l}; \dots ;\tau_{P_R}^{l}]$. With $\tau_{\mathcal{P}}^{l}$, the CTS Attention replace \cref{eq:MMA} as follow:
\begin{equation}
    \label{eq:contexual_sharing}
    \resizebox{0.89\linewidth}{!}{$\text{A}^{\text{CTS}}(\tau_{I_o}^{l},\tau_{P_o}^{l},\tau_{\mathcal{P}}^{l},M) = \text{softmax}\big((QK^T+M)/\sqrt{d}\big)V$},
\end{equation}
where $Q=[Q_{I_o};Q_{P_o}]$, $K=[K_{I_o};K_{P_o};{K_{\mathcal{P}}}]$ and $V=[V_{I_{P}};V_{P_o};V_{\mathcal{P}}]$. By concatenating contextual tokens instead of vision tokens, we solve the main restriction of the share attention module — the computational overheads, R references require $(2R-1)$ times computation and memory sources for attention operation, while concatenating contextual tokens only require $((R-1)+(n_{P}/n_{I})R)$ nearly half times of sources ($n_{P}/n_{I}=0.08$ for SD3~\cite{sd3}), allowing us to incorporate more references efficiently. 

\subsubsection{Reference Contextual Masking (RCM)}
\label{sec:RCM}
Along with the increase of references, the conflict of references grows severe. For instance, in \cref{fig:teaser}, we aim to generate the background of a starry night instead of the jungle background from the \textit{``dinosaur''} reference or the dark night background from the \textit{``breathing fire''} reference. Thus, we design \textbf{References Contextual Masking} (RCM) to select only the vision tokens we need from the given reference instead of the whole image. 

Inspired by prior works~\cite{diffedit, attend_and_excite, ledits++} that show cross-attention maps can already play a similar function as semantic matching, we extract the semantic connection between the instruction and each visual token. For the visual tokens $\tau_{I_r}^{l}$ from the given reference $I_r$, we utilize $\tau_{P_o}^{l}$ to measure the corresponding semantic connection of each token by computing the attention score. We then compare the summation of the attention scores between the query of $\tau_{I_r}^{l}$ and the key of $\tau_{P_o}^{l}$ on each visual token and retain only the salient tokens via a binarization operation. The resulting contextual mask for $i^{th}$ vision token of $r^{th}$ reference is:
\begin{equation}
\resizebox{0.89\linewidth}{!}{$M_{i,r}^{\text{RCM}} = \text{BI} \Bigg( \sum_{j=1}^{n_{P}} \text{softmax}\big(Q_{I_r}^{l} [K_{I_r}^{l};K_{P_o}^{l}]^T\big)_{i, n_I + j - 1}\Bigg)$},
\end{equation}
where ``BI'' denotes the binarize algorithm. Specifically, we utilized Otsu Algorithm~\cite{otsu} to detach the foreground from the background. We gather $M_{i,r}^{\text{RCM}}$ for each vision token to form the reference contextual mask $M_{r}^{\text{RCM}}$ as an attention mask during the forward process of $\tau_{I_r}^{l}$ and $\tau_{P_r}^{l}$:
\begin{equation}
    [\tau_{I_r}^{l+1};\tau_{P_r}^{l+1}] = [\tau_{I_r}^{l};\tau_{P_r}^{l}] + \text{A}(\tau_{I_r}^{l},\tau_{P_r}^{l},M_{r}^{\text{RCM}}),
\end{equation}
as shown in \cref{fig:pipline}. By summing and binarizing the attention score into a mask, we can restrict the contextual tokens to only learn contextual information from specific vision tokens, reducing redundant information from reference.

\subsubsection{Winner Takes All (WTA)}
\label{sec:WTA}
As a training-free modification applied to a pre-trained model, our approach inevitably faces distribution shifts, leading to unsatisfactory generation results, as suggested in \cite{style_injection,zstar,freeenhance}. 

To address this, we propose a novel \textbf{Winner Takes All} (WTA) strategy to mitigate conflicts between multiple references. We assume that, for each vision token, only one or two references are needed at a time. For instance, background references are not necessary at every step and layer when generating an object. Leveraging the semantic correlation capability of our T2I model, we first measure the attention score of references to a given vision token and selectively keep only the contextual tokens with highest attention score to minimize distribution disturbances. This is formulated as:
\begin{equation}
s_{i,r} = \sum_{j=1}^{n_{P}} \text{softmax}\big(Q_{I_o} K_{P_r}^T\big)_{i,j}.
\end{equation}
To perform the winner-takes-all criteria, we set $r^{*}_{i} = \arg\max_r s_{i,r}$. As such, the WTA mask is applied as:
\begin{small}
\begin{equation}
M^{\text{WTA}}_{i,j} =
\begin{cases}
-\infty, & (n_{I}+n_{P})<j \leq n_{I} + n_{P}\cdot r^*_i \\ 
-\infty, & n_{I} + n_{P}+n_{P}\cdot r^*_i \leq j \\
0, & \text{otherwise}
\end{cases},
\end{equation}
\end{small}
where $i$ denotes the $i^{th}$ vision token. We then incorporate WTA mask $M^{\text{WTA}}$ into contextual sharing \cref{eq:contexual_sharing}:
\begin{equation}
    \resizebox{0.86\linewidth}{!}{$[\tau_{I_o}^{l+1};\tau_{P_o}^{l+1}] = [\tau_{I_o}^{l};\tau_{P_o}^{l}] + \text{A}^{\text{CTS}}(\tau_{I_o}^{l},\tau_{P_o}^{l}, \tau_{\mathcal{P}}^{l},M^{\text{WTA}})$}.
\end{equation}
The illustration can be found in the upper part of \cref{fig:pipline}. The WTA module may assign different contextual tokens to vision tokens across different layers, effectively mixing multiple references.
\section{Experiment}
\label{sec:exp}
As a unified TI2I framework, our proposed TF-TI2I shares similarities with the well-known Emu2~\cite{emu2} and supports various downstream tasks, including object customization~\cite{textual_inversion,dreambooth,suti,re_imagen,kosmosg,omnigen}, consistent synthesis~\cite{masactrl,style_align}, style transfer~\cite{style_injection,stylebooth,zstar,tfgph} and image editing~\cite{sdedit,diffedit,p2p,pnp,tis}. To evaluate its customization and editing capabilities, we use DreamBench~\cite{dreambooth} (750 instances) and Wild-TI2I~\cite{pnp} (148 instances). Additionally, we introduce FG-TI2I for fine-grained evaluation of the TI2I task. Since our evaluation involves both textual and visual inputs, we adopt a diverse set of metrics for a comprehensive assessment:  

\begin{itemize}
    \item \textbf{Prompt following}: CLIP-Text~\cite{clip} (CP), CLIP-Directional Score~\cite{clip_dir} (CDS), Image Reward~\cite{imagereward} (IR), Human Preference Score~\cite{hpsv2} (HPS)
    \item \textbf{Reference alignment}: CLIP-Image~\cite{clip} (CP-I), DINO Similarity~\cite{dinov2} (DI), LPIPS~\cite{lpips} (LP)
    \item \textbf{Generation quality}: Aesthetic Score~\cite{sd} (AS), Image Reward~\cite{imagereward} (IR), Human Preference Score~\cite{hpsv2} (HPS)
\end{itemize}

\begin{figure}
    \centering
    \includegraphics[width=1\linewidth]{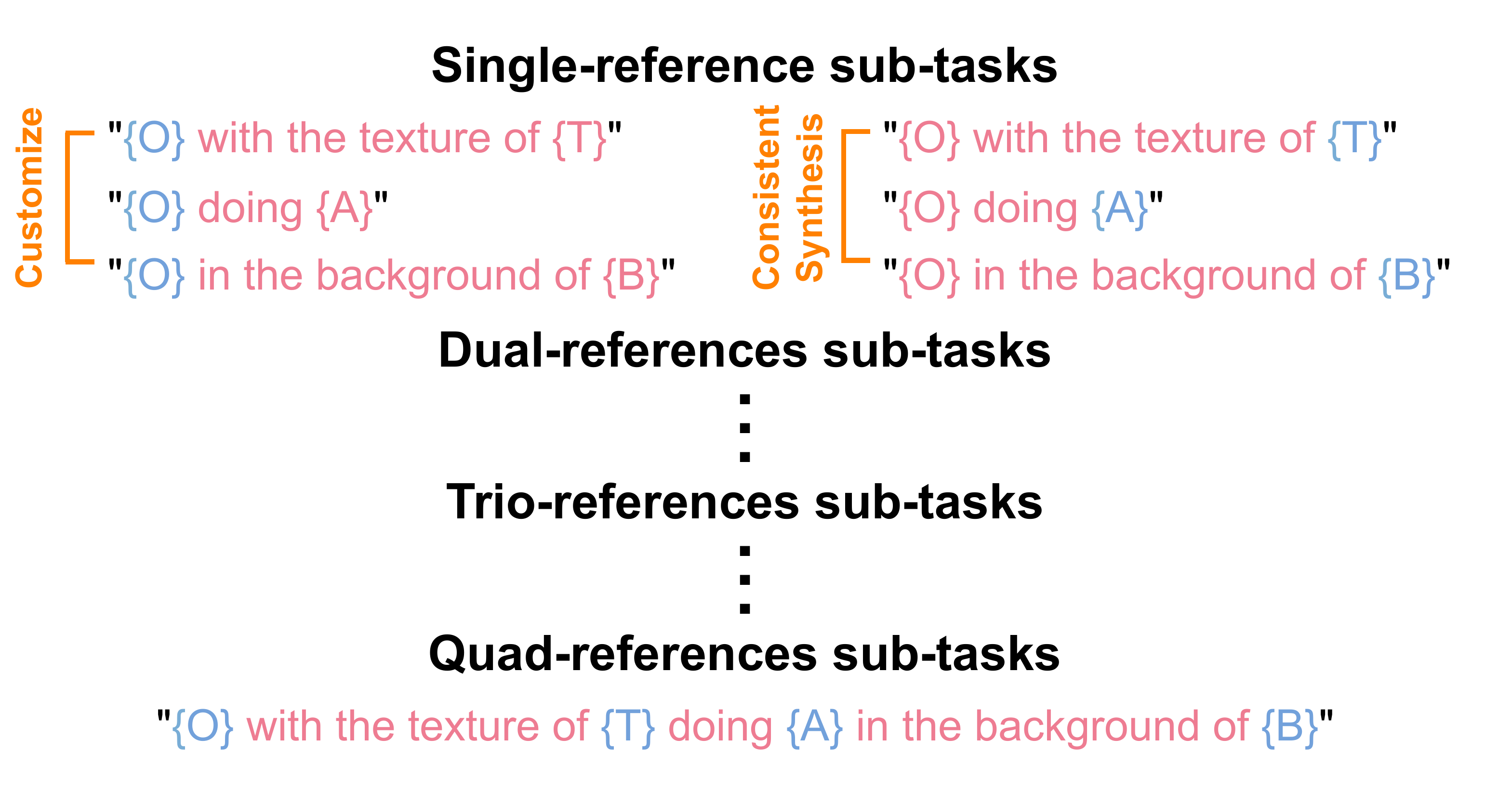}
    \caption{Illustration of sub-tasks in FG-TI2I, where we abbreviate Object, Texture, Action, and Background into O, T, A, and B, respectively. The text-only input is denoted with \textcolor{text_red}{red}, and image-support input is denoted with \textcolor{image_blue}{blue}. The input sub-tasks are categorized by the number of image references.}
    \label{fig:fg_ti2i}
\end{figure}
\subsection{Fine-Grained TI2I Benchmark}
Although benchmarks~\cite{p2p,dreambooth,editbench,tf_icon,tfgph,brushnet} exist for evaluating single aspect of image input, such as conditioning on object, background, and style, the evaluation of unified TI2I—where a single instruction incorporates multiple aspects of image references—remains relatively unexplored. To address this gap, we propose the Fine-Grained TI2I (FG-TI2I) benchmark, the first image generation benchmark that integrates multiple reference aspects within a single instruction. FG-TI2I adopts a unified representation for common image generation tasks, utilizing a single instruction template with four aspects of reference: \textit{``\{Object\} with the texture of \{Texture\} doing \{Action\} in the background of \{Background\}''}. As illustrated in \cref{fig:fg_ti2i}, we categorize the sub-tasks based on the number of image references. For each sub-task, we generate 100 prompts using ChatGPT-4o and create the corresponding image references using Flux 1.0 dev~\cite{flux}. By introducing image references at different aspects, we can systematically evaluate a model's capability in various downstream tasks.

\begin{table*}[h]
 \scriptsize
    \centering
    \begin{tabular}{c  c c c  c c c  c c c  c c c  c c c  c c c}
        \toprule

        & \multicolumn{3}{c}{\textcolor{image_blue}{OBJ} x \textcolor{text_red}{TEX}}   
        & \multicolumn{3}{c}{\textcolor{image_blue}{OBJ} x \textcolor{text_red}{ACT}}  
        & \multicolumn{3}{c}{\textcolor{image_blue}{OBJ} x \textcolor{text_red}{BG}}  
        & \multicolumn{3}{c}{\textcolor{text_red}{OBJ} x \textcolor{image_blue}{TEX}} 
        & \multicolumn{3}{c}{\textcolor{text_red}{OBJ} x \textcolor{image_blue}{ACT}}  
        & \multicolumn{3}{c}{\textcolor{text_red}{OBJ} x \textcolor{image_blue}{BG}}  
        \\

        \cmidrule(lr){2-4}\cmidrule(lr){5-7}\cmidrule(lr){8-10}\cmidrule(lr){11-13}\cmidrule(lr){14-16}\cmidrule(lr){17-19}
        & CP & DI & IR & CP & DI & IR & CP & DI & IR & CP & DI & IR & CP & DI & IR & CP & DI  & IR \\
        \cmidrule(lr){1-19}
        MasaCtrl~\cite{masactrl}    & 26.8 & \textbf{96.4} & -0.26 & 25.4 & \textbf{96.5} & -0.92 & 26.6 & \textbf{95.7} & -0.31 & 26.5 & \textbf{91.7} & 0.10 & 27.0 & \textbf{88.7} & 0.09 & 26.9 & \textbf{97.3} & 0.14 \\
        StyleAlign~\cite{style_align}          & 27.3 & \underline{94.3} & 0.10 & 28.3 & \underline{92.9} & 0.28 & 28.3 & \underline{89.9} & 0.56 & 27.7 & \underline{88.9} & 0.33 & \underline{29.3} & \underline{87.7} & \underline{0.88} & \underline{29.5} & \underline{89.6} & 0.82 \\
        OmniGen~\cite{omnigen} & \underline{28.2} & 84.8 & \underline{0.36} & \underline{28.6} & 82.7 & \underline{0.62} & \underline{29.2} & 81.2 & \underline{1.09} & 28.4 & 75.8 & \underline{0.40} & 28.7 & 76.1 & 0.85 & 29.2 & 85.5 & \underline{1.12} \\
        Emu2~\cite{emu2}   & \underline{28.2} & 71.2 & 0.10 & 27.5 & 75.5 & 0.34 & 28.2 & 72.0 & 0.66 & \underline{28.5} & 63.8 & 0.10 & 27.6 & 62.9 & 0.36 & 28.2 & 75.7 & 0.68 \\
        Ours        & \textbf{29.3} & 87.5 & \textbf{0.70} & \textbf{30.3} & 87.1 & \textbf{1.21} & \textbf{29.3} & 84.1 & \textbf{1.14} & \textbf{30.1} & 72.8 & \textbf{0.67} & \textbf{30.7} & 75.8 & \textbf{1.30} & \textbf{30.1} & 84.4 & \textbf{1.26} \\
        \bottomrule
    \end{tabular}
    \caption{Quantitative comparison over FG-TI2I single-entry, where we respectively abbreviate Object, Texture, Action, and Background into OBJ, TEX, ACT, and BG. \textbf{Bold} highlights the best result, and \underline{underlines} mark the second-best.}
    \label{tab:main}
\end{table*}

\floatsetup[table]{capposition=bottom}
\newfloatcommand{capbtabbox}{table}[][\FBwidth]

\begin{table}
\begin{floatrow}
\capbtabbox{
\scriptsize
\begin{tabular}{ccc}
\toprule
  & CP &CP-I  \\
\cmidrule(lr){1-3}
 Tex-Inv~\cite{textual_inversion} & 25.5& 78.0\\
 DB~\cite{dreambooth} & 30.5& 80.3\\
 ReImagen~\cite{re_imagen} & 27.0& 74.0\\
 SuTI~\cite{suti} & 30.4& 81.9\\
 Kosmos-G~\cite{kosmosg} & 28.7& \textbf{84.7}\\
 OmniGen~\cite{omnigen} & 31.5& 80.1\\
 Ours & \textbf{33.1} & 79.1\\
\bottomrule
\end{tabular}
}{
 \caption{Quantitative comparison on DreamBench.}
 \label{Tab:dreambench}
}
\hspace{-0.7cm}
\capbtabbox{
\scriptsize
\begin{tabular}{ccc}
\toprule
  & CP & CDS  \\
\cmidrule(lr){1-3}
 SDEdit~\cite{sdedit} & 27.5 &0.122 \\
 DiffEdit~\cite{diffedit} & 26.3 & 0.088\\
 P2P~\cite{p2p} & 28.4 & 0.193\\
 PnP~\cite{pnp} & 28.5 & 0.202\\
 MasaCtrl~\cite{masactrl} & 29.3 & 0.209\\
 TIS~\cite{tis} & 29.0 & \textbf{0.210}\\
 Ours & \textbf{30.4} & 0.115 \\
\bottomrule
\end{tabular}
}{
\scriptsize
\caption{Quantitative comparison
on Wild-TI2I}
\label{Tab:wild_ti2i}
}
\end{floatrow}
\end{table}

\begin{figure*}
    \centering
    \includegraphics[width=1\linewidth]{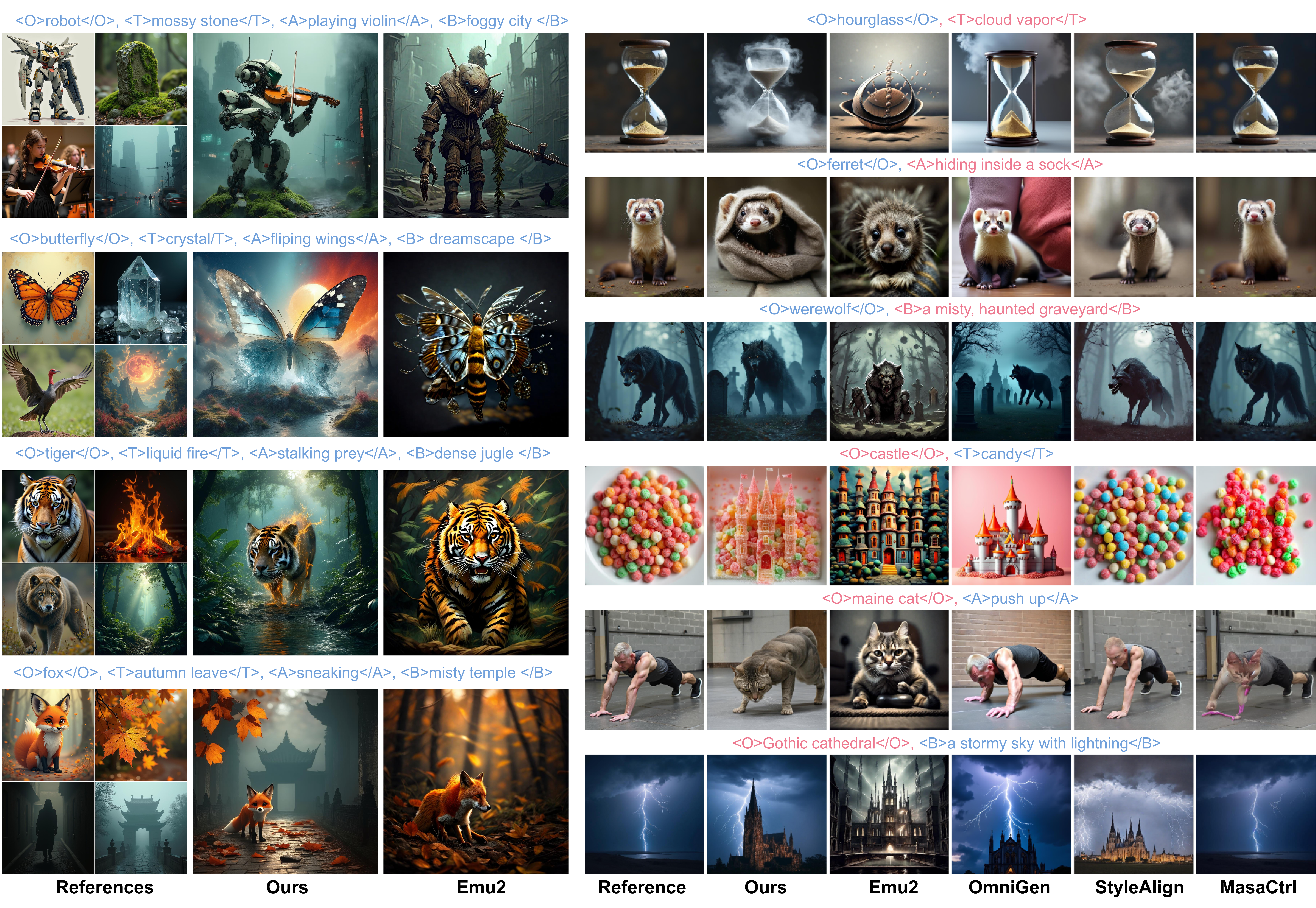}
    \caption{Qualitative comparison of Quad-references sub-tasks (left) and Single-reference sub-tasks (right) in FG-TI2I. The input Object, Texture, Action, and Background—are denoted as O, T, A, and B. We use \textcolor{text_red}{red} for text-only input and \textcolor{image_blue}{blue} for reference-supported input.}
    \label{fig:main_qualitative}
\end{figure*}

\begin{figure}[htbp]
\footnotesize
    \centering
    \begin{subfigure}[b]{0.45\linewidth}{\includegraphics[width=\linewidth]{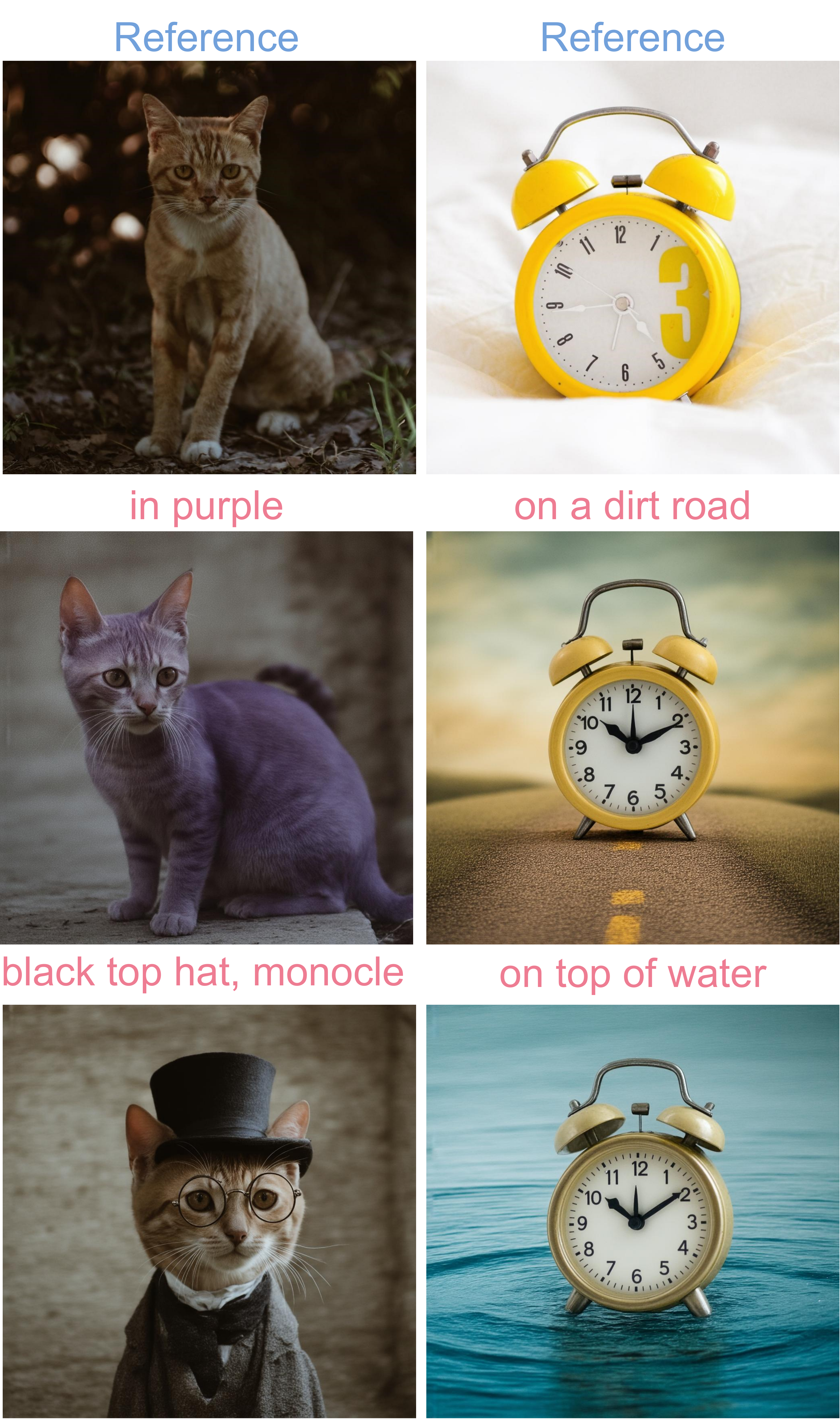}}
    \caption{DreamBench~\cite{dreambooth}}
    \label{fig:db}
    
    \end{subfigure}
    \begin{subfigure}[b]{0.45\linewidth}{\includegraphics[width=\linewidth]{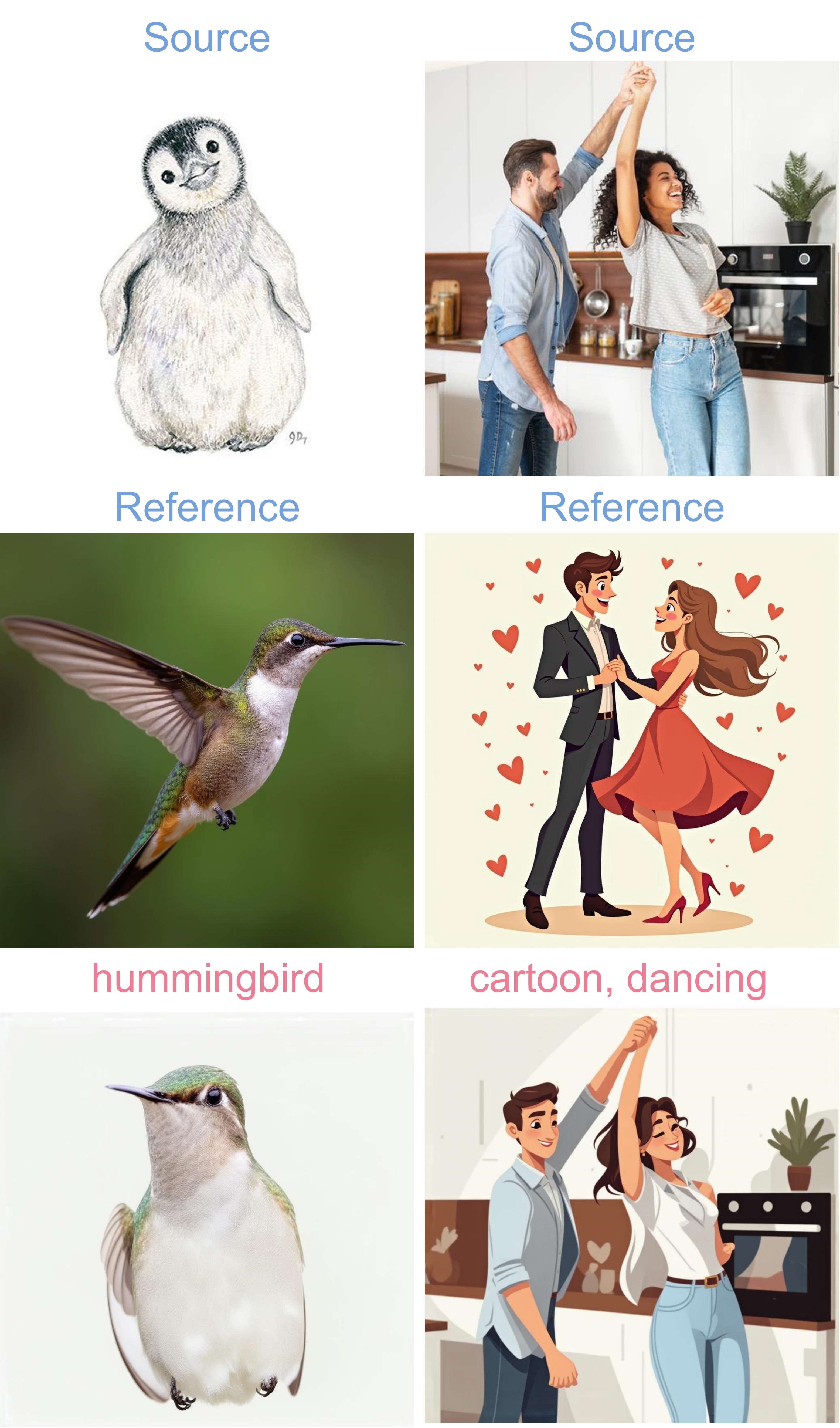}}
    \caption{Wild-TI2I~\cite{pnp}}
    \label{fig:wt}
    \end{subfigure}
\caption{Qualitative results of TF-TI2I on Customization~\cite{dreambooth} and Editing~\cite{pnp}. For Wild-TI2I, we first generate a reference for each sample to support text-only editing instructions.}
\label{fig:db_wt}
\end{figure}

\subsection{Qualitative Results}
For multiple-reference test cases in \cref{fig:main_qualitative}, TF-TI2I achieves better compliance with all given references and higher image quality compared to the previous TI2I method, Emu2~\cite{emu2}. For instance, in row 1, although Emu2 produces a reference-like background, the other input components, \ie, the given object, texture, and action, are missing in the output. Similarly, while Emu2 successfully generates the given object reference in cases 2, 3, and 4, the texture, action, and background are not preserved in the generated results. These findings confirm a key limitation of Emu2: constrained by its training design, Emu2 excels in object customization but struggles to generalize well to the more creative and complicated TI2I instructions introduced in our proposed FG-TI2I benchmark.

Notably, even when compared to OmniGen~\cite{omnigen}, StyleAlign~\cite{style_align}, and MasaCtrl~\cite{masactrl}, which are designed for single-reference generation, our proposed method still demonstrates superior performance, especially in texture-related and action-related prompts. This is achieved through our proposed contextual token-sharing mechanism, which enables TF-TI2I to achieve a better balance between textual instructions and image references by leveraging image references as textual tokens. As a result, TF-TI2I generates more texture-aligned outputs, as shown in rows 1 and 4 on the right side of~\cref{fig:main_qualitative}, successfully synthesizing the cloud vapor hourglass based on textual instructions and the candy castle based on visual references. Additionally, in rows 2 and 5, TF-TI2I better aligns with action-related instructions, effectively following both textual and visual guidance.

Furthermore, TF-TI2I is readily compatible with customization and editing tasks without requiring any modifications, as shown in \cref{fig:db_wt}. It seamlessly enables object modifications (\ie, changing the color and outfit of a cat) and object transfer to different backgrounds. 

\subsection{Quantitative Results}
Our quantitative comparison in \cref{tab:main} primarily focuses on single-reference evaluations, as most baseline methods are designed for such settings. Our proposed TF-TI2I framework demonstrates superior performance in both prompt-following and image quality metrics. This advantage stems from our proposed CTS strategy, which integrates additional contextual tokens into the generation pipeline. These contextual tokens, functionally similar to textual tokens but enriched with visual information, introduce minimal disturbance to the original T2I model. Consequently, TF-TI2I achieves higher prompt alignment and better image quality. 

However, as a trade-off, our method exhibits slightly lower performance in DINO Similarity (DI) compared to training-based approaches like OmniGen~\cite{omnigen} and inversion-based methods such as StyleAlign~\cite{style_align} and MasaCtrl~\cite{masactrl}. This limitation arises because TF-TI2I is not designed for direct reference copying but instead encodes references into high-level semantics, thereby sacrificing low-level details but ensuring instruction compliance.

A similar trend can be observed in the reported performance in \cref{Tab:dreambench} and \cref{Tab:wild_ti2i} with TF-TI2I achieves state-of-the-art (SOTA) results in CLIP-Text (CP) but does not surpass all the existing methods in terms of CLIP-Image (CP-I) and CLIP-Directional Score (CDS) as our approach might alter the reference to make the output compatible with the new scene. Besides, TF-TI2I maintains competitive reference-alignment performance—outperforming the TI2I model EMU2 across all FG-TI2I sub-tasks, surpassing Textual Inversion~\cite{textual_inversion} by 1\%, and trailing the SOTA customization method OmniGen by only 7\%.

\subsection{Ablation study}
\begin{figure}[htbp]
\footnotesize
    \centering
    \begin{subfigure}[b]{0.32\linewidth}{\includegraphics[width=\linewidth]{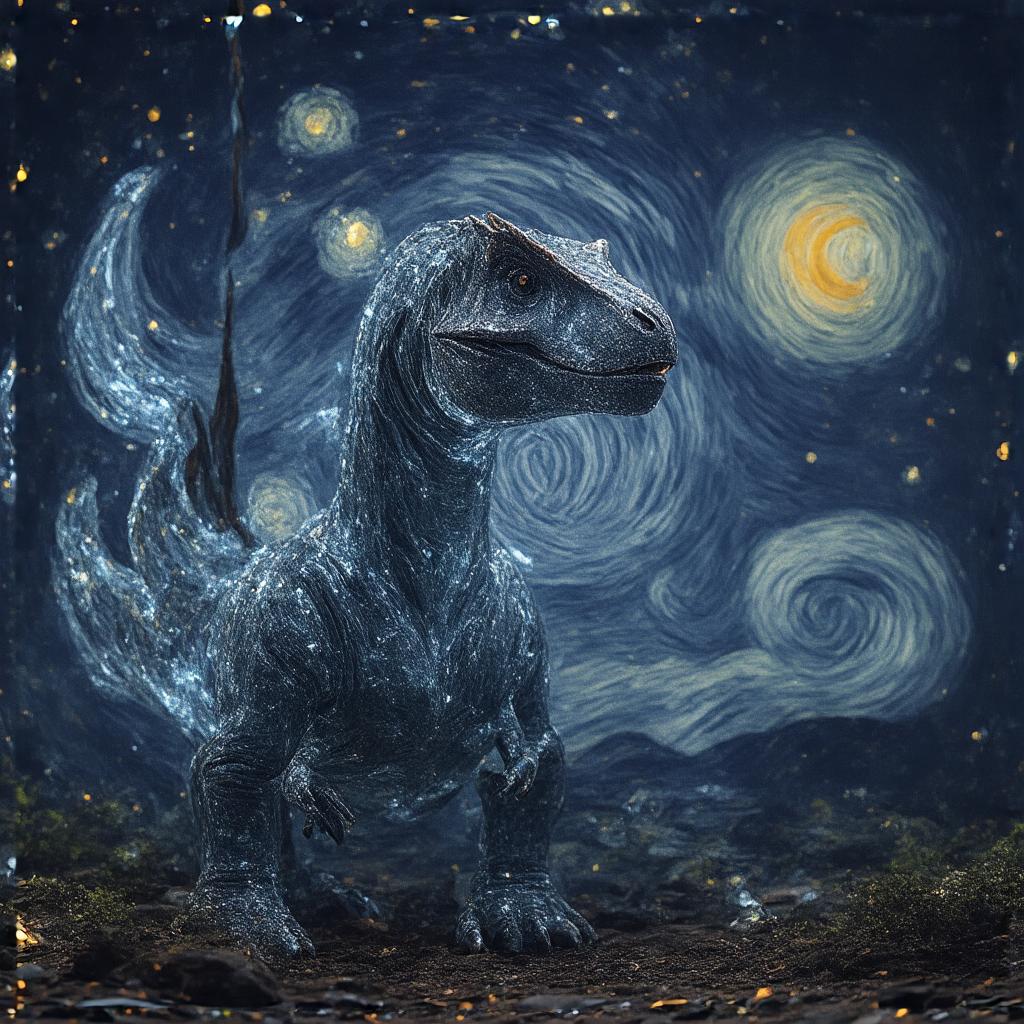}}
    \caption{CTS only}
    \label{abla:cts}
    \end{subfigure}
    \begin{subfigure}[b]{0.32\linewidth}{\includegraphics[width=\linewidth]{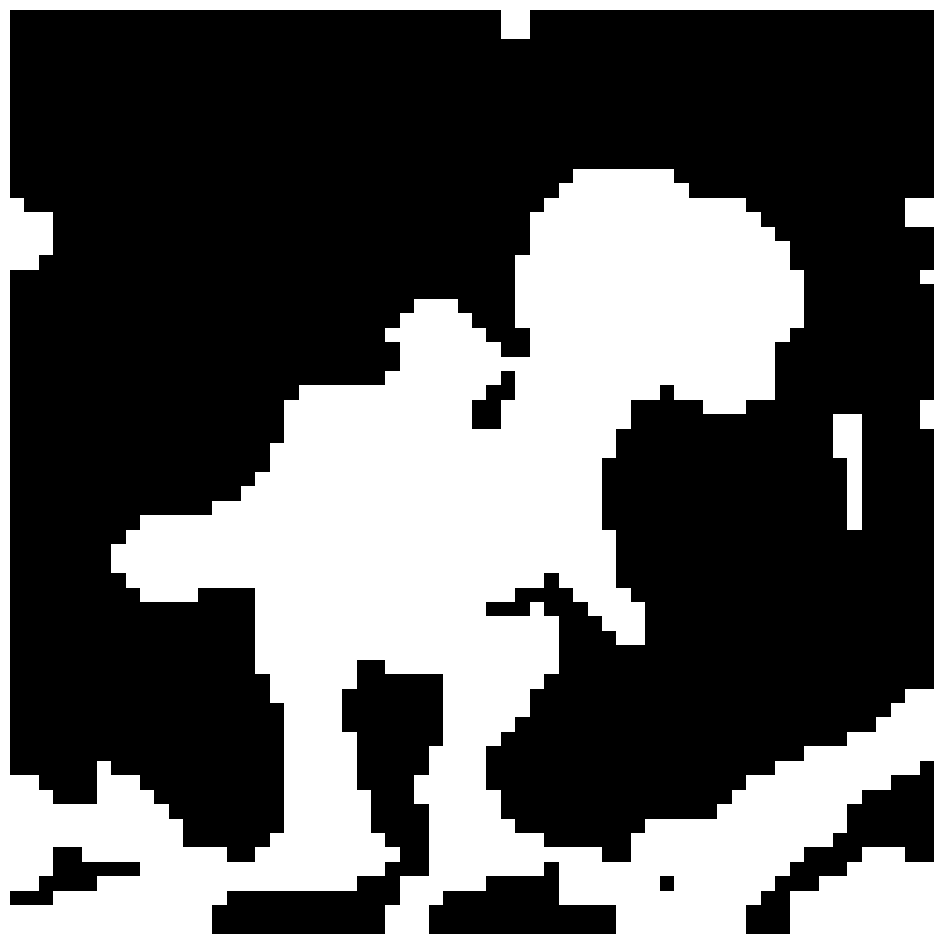}}
    \caption{$M^\text{RCM}_1$}
    \label{abla:M_R}
    \end{subfigure}
        \begin{subfigure}[b]{0.32\linewidth}{\includegraphics[width=\linewidth]{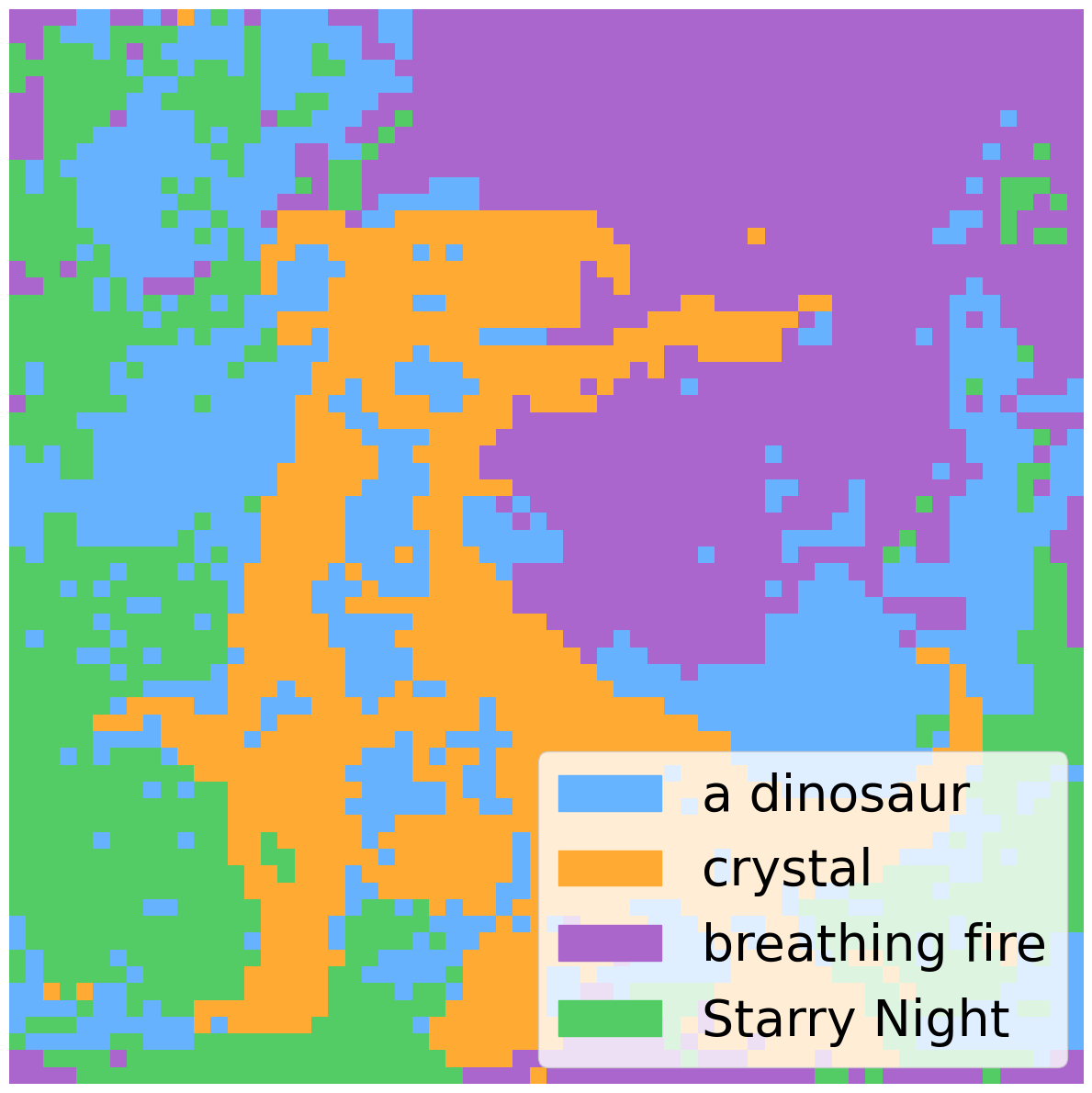}}
    \caption{$M^\text{WTA}_o$}
    \label{abla:M_W}
    \end{subfigure}

    \begin{subfigure}[b]{0.32\linewidth}{\includegraphics[width=\linewidth]{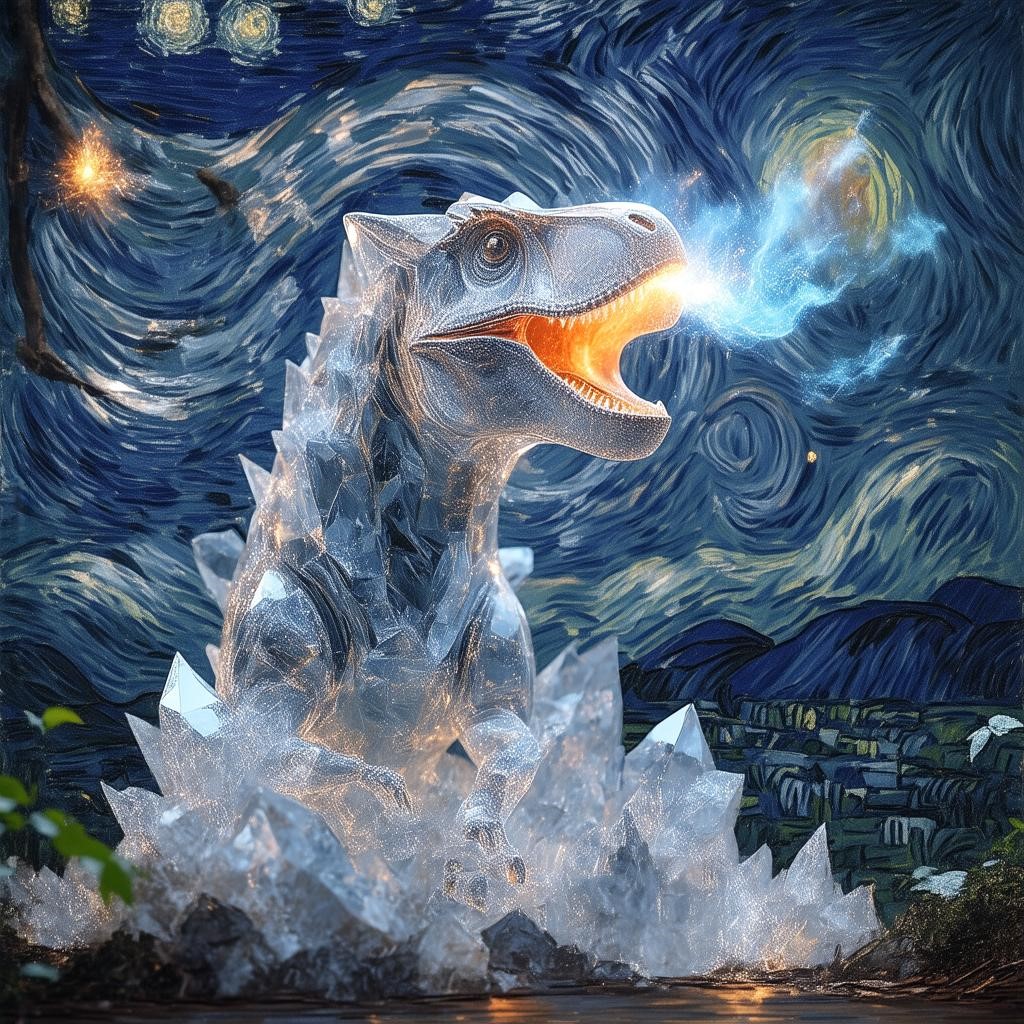}}
    \caption{+RCM+WTA}
    \label{abla:rcm_wta}
    \end{subfigure}
    \begin{subfigure}[b]{0.32\linewidth}{\includegraphics[width=\linewidth]{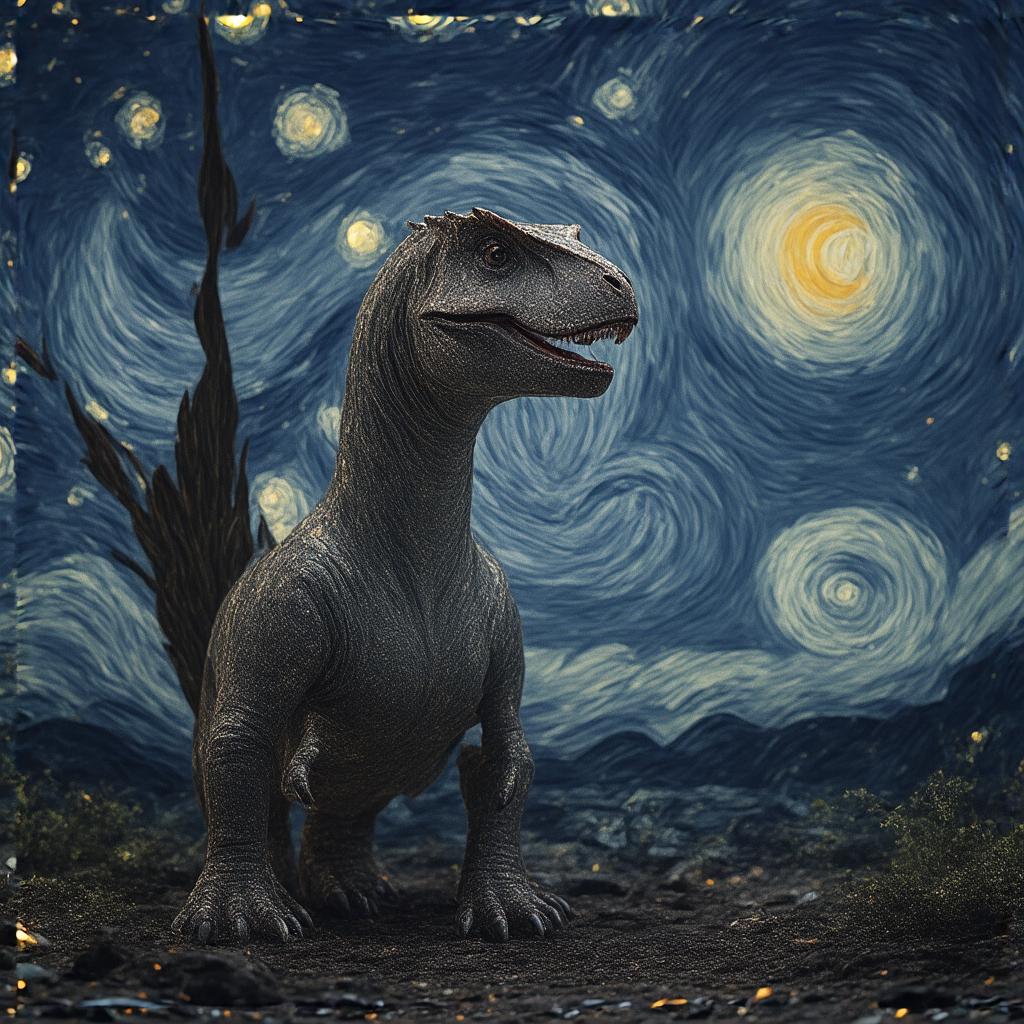}}
    \caption{+RCM}
    \label{abla:rcm}
    \end{subfigure}
        \begin{subfigure}[b]{0.32\linewidth}{\includegraphics[width=\linewidth]{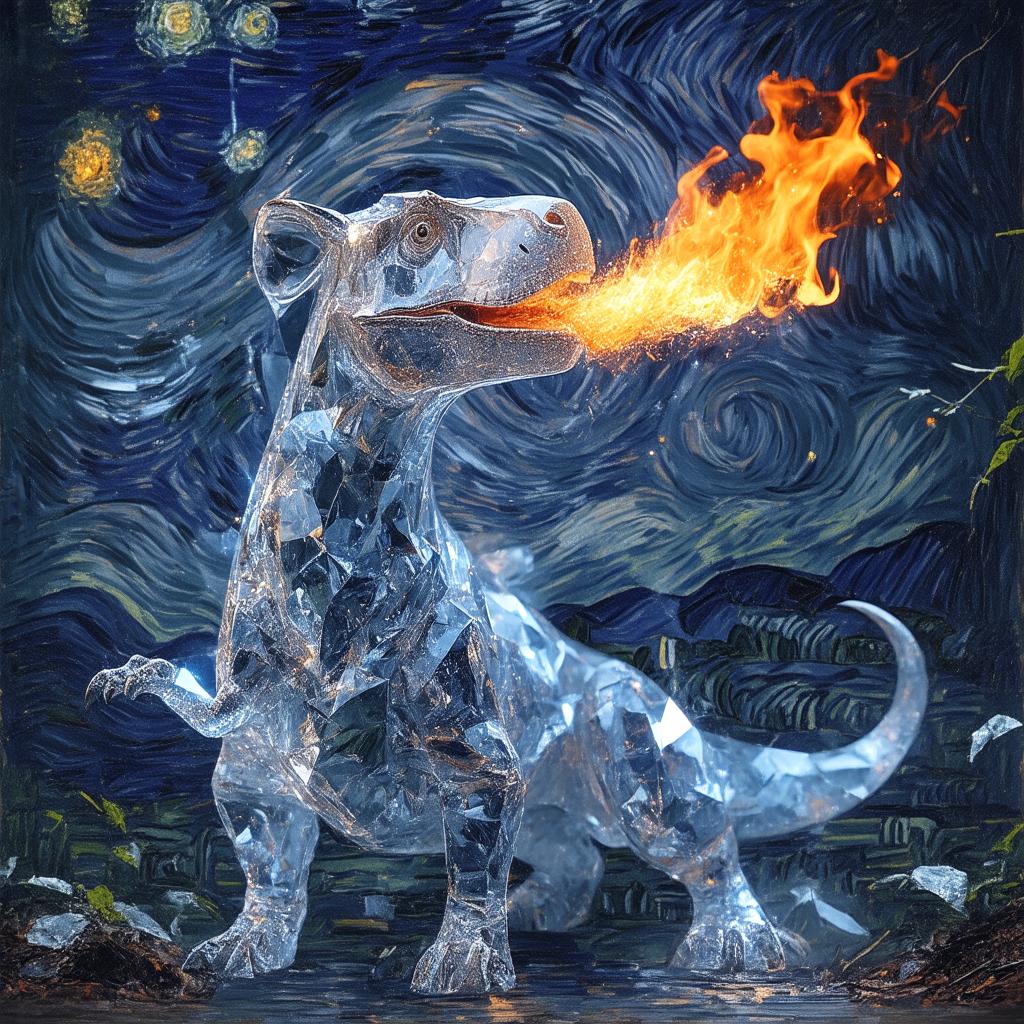}}
    \caption{+WTA}
    \label{abla:wta}
    \end{subfigure}
\caption{Ablation study of TF-TI2I, using \cref{fig:teaser} as example.}
\end{figure}

\noindent \textbf{RCM.} By restricting the learning of contextual tokens to only instruction-related vision tokens using $M^\text{RCM}_r$, we can ease mutual interference between different references, as shown in \cref{abla:M_R}, where $M^\text{RCM}_1$ is the contextual mask for \textit{the dinosaur}. Comparing \cref{abla:rcm} to \cref{abla:cts}, both the generated background \textit{Starry Night} and the generated object \textit{a dinosaur} exhibit greater similarity to their respective input references, demonstrating the effectiveness of RCM in reducing mutual disturbance between references.

\noindent \textbf{WTA.} Designed to mitigate distribution shifts caused by additional reference inputs, it achieves this by measuring the saliency score of each reference and generating corresponding reference assignments for each vision token, as shown in \cref{abla:M_W}. This strategy ensures both reference adherence and visually satisfactory results. As demonstrated in \cref{abla:rcm_wta} and \cref{abla:wta}, the generated images exhibit richer details and colors compared to \cref{abla:cts} and \cref{abla:rcm}.

\section{Conclusion}
In this paper, we investigate the implicit-context learning capability of MM-DiT and leverage this insight to develop TF-TI2I, a training-free text-and-image-to-image generation pipeline. Our approach surpasses previous methods in effectively integrating multiple types of image references. To overcome the limitations of existing benchmarks, we introduce FG-TI2I, a more comprehensive evaluation framework for TI2I tasks. Experimental results across various benchmarks demonstrate the effectiveness of our method. However, mutual interference between image references remains significant due to the limited precision of RCM, as MM-DiT is not explicitly optimized for semantic segmentation. In future work, we aim to mitigate this interference to achieve finer control over generation.
\label{sec:conclusion}

\section*{Acknowledgments}
  This work is partially supported by the National Science and Technology Council, Taiwan under Grants NSTC-112-2221-E-A49-059-MY3 and NSTC-112-2221-E-A49-094-MY3.
{
    \small
    \bibliographystyle{ieeenat_fullname}
    \bibliography{main}

@String(TOG= {ACM Trans. Graph.})

@String(TOG   = {ACM TOG})

@inproceedings{sd3,
  title={Scaling rectified flow transformers for high-resolution image synthesis},
  author={Esser, Patrick and Kulal, Sumith and Blattmann, Andreas and Entezari, Rahim and M{\"u}ller, Jonas and Saini, Harry and Levi, Yam and Lorenz, Dominik and Sauer, Axel and Boesel, Frederic and others},
  booktitle={Forty-first International Conference on Machine Learning},
  year={2024}
}

@misc{sd,
      title={High-Resolution Image Synthesis with Latent Diffusion Models}, 
      author={Robin Rombach and Andreas Blattmann and Dominik Lorenz and Patrick Esser and Björn Ommer},
      year={2022},
      eprint={2112.10752},
      archivePrefix={arXiv},
      primaryClass={cs.CV},
      url={https://arxiv.org/abs/2112.10752}, 
}

@misc{controlnet,
      title={Adding Conditional Control to Text-to-Image Diffusion Models}, 
      author={Lvmin Zhang and Anyi Rao and Maneesh Agrawala},
      year={2023},
      eprint={2302.05543},
      archivePrefix={arXiv},
      primaryClass={cs.CV},
      url={https://arxiv.org/abs/2302.05543}, 
}

@article{uni_controlnet,
  title={Uni-controlnet: All-in-one control to text-to-image diffusion models},
  author={Zhao, Shihao and Chen, Dongdong and Chen, Yen-Chun and Bao, Jianmin and Hao, Shaozhe and Yuan, Lu and Wong, Kwan-Yee K},
  journal={Advances in Neural Information Processing Systems},
  volume={36},
  year={2024}
}

@inproceedings{brushnet,
  title={Brushnet: A plug-and-play image inpainting model with decomposed dual-branch diffusion},
  author={Ju, Xuan and Liu, Xian and Wang, Xintao and Bian, Yuxuan and Shan, Ying and Xu, Qiang},
  booktitle={European Conference on Computer Vision},
  pages={150--168},
  year={2024},
  organization={Springer}
}

@inproceedings{powerpaint,
  title={A task is worth one word: Learning with task prompts for high-quality versatile image inpainting},
  author={Zhuang, Junhao and Zeng, Yanhong and Liu, Wenran and Yuan, Chun and Chen, Kai},
  booktitle={European Conference on Computer Vision},
  pages={195--211},
  year={2024},
  organization={Springer}
}

@inproceedings{style_injection,
  title={Style injection in diffusion: A training-free approach for adapting large-scale diffusion models for style transfer},
  author={Chung, Jiwoo and Hyun, Sangeek and Heo, Jae-Pil},
  booktitle={Proceedings of the IEEE/CVF Conference on Computer Vision and Pattern Recognition},
  pages={8795--8805},
  year={2024}
}

@misc{tfgph,
      title={Training-and-Prompt-Free General Painterly Harmonization via Zero-Shot Disentenglement on Style and Content References}, 
      author={Teng-Fang Hsiao and Bo-Kai Ruan and Hong-Han Shuai},
      year={2024},
      eprint={2404.12900},
      archivePrefix={arXiv},
      primaryClass={cs.CV},
      url={https://arxiv.org/abs/2404.12900}, 
}

@misc{sdxl,
      title={SDXL: Improving Latent Diffusion Models for High-Resolution Image Synthesis}, 
      author={Dustin Podell and Zion English and Kyle Lacey and Andreas Blattmann and Tim Dockhorn and Jonas Müller and Joe Penna and Robin Rombach},
      year={2023},
      eprint={2307.01952},
      archivePrefix={arXiv},
      primaryClass={cs.CV},
      url={https://arxiv.org/abs/2307.01952}, 
}

@misc{flux,
    author={Black Forest Labs},
    title={FLUX},
    year={2024},
    howpublished={\url{https://github.com/black-forest-labs/flux}},
}

@inproceedings{sana,
title={{SANA}: Efficient High-Resolution Text-to-Image Synthesis with Linear Diffusion Transformers},
author={Enze Xie and Junsong Chen and Junyu Chen and Han Cai and Haotian Tang and Yujun Lin and Zhekai Zhang and Muyang Li and Ligeng Zhu and Yao Lu and Song Han},
booktitle={The Thirteenth International Conference on Learning Representations},
year={2025},
url={https://openreview.net/forum?id=N8Oj1XhtYZ}
}

@misc{janusflow,
      title={JanusFlow: Harmonizing Autoregression and Rectified Flow for Unified Multimodal Understanding and Generation}, 
      author={Yiyang Ma and Xingchao Liu and Xiaokang Chen and Wen Liu and Chengyue Wu and Zhiyu Wu and Zizheng Pan and Zhenda Xie and Haowei Zhang and Xingkai yu and Liang Zhao and Yisong Wang and Jiaying Liu and Chong Ruan},
      journal={arXiv preprint arXiv:2411.07975},
      year={2024}
}

@inproceedings{smartcontrol,
  title={Smartcontrol: Enhancing controlnet for handling rough visual conditions},
  author={Liu, Xiaoyu and Wei, Yuxiang and Liu, Ming and Lin, Xianhui and Ren, Peiran and Xie, Xuansong and Zuo, Wangmeng},
  booktitle={European Conference on Computer Vision},
  pages={1--17},
  year={2024},
  organization={Springer}
}

@inproceedings{anydoor,
  title={Anydoor: Zero-shot object-level image customization},
  author={Chen, Xi and Huang, Lianghua and Liu, Yu and Shen, Yujun and Zhao, Deli and Zhao, Hengshuang},
  booktitle={Proceedings of the IEEE/CVF Conference on Computer Vision and Pattern Recognition},
  pages={6593--6602},
  year={2024}
}

@article{freecond,
  title={FreeCond: Free Lunch in the Input Conditions of Text-Guided Inpainting},
  author={Hsiao, Teng-Fang and Ruan, Bo-Kai and Tsai, Sung-Lin and Wu, Yi-Lun and Shuai, Hong-Han},
  journal={arXiv preprint arXiv:2412.00427},
  year={2024}
}

@inproceedings{smartbrush,
  title={Smartbrush: Text and shape guided object inpainting with diffusion model},
  author={Xie, Shaoan and Zhang, Zhifei and Lin, Zhe and Hinz, Tobias and Zhang, Kun},
  booktitle={Proceedings of the IEEE/CVF Conference on Computer Vision and Pattern Recognition},
  pages={22428--22437},
  year={2023}
}

@article{zstar,
  title={Z-star: Zero-shot Style Transfer via Attention Rearrangement},
  author={Deng, Yingying and He, Xiangyu and Tang, Fan and Dong, Weiming},
  journal={arXiv preprint arXiv:2311.16491},
  year={2023}
}

@article{stylebooth,
  title={StyleBooth: Image Style Editing with Multimodal Instruction},
  author={Han, Zhen and Mao, Chaojie and Jiang, Zeyinzi and Pan, Yulin and Zhang, Jingfeng},
  journal={arXiv preprint arXiv:2404.12154},
  year={2024}
}

@inproceedings{dreambooth,
  title={Dreambooth: Fine tuning text-to-image diffusion models for subject-driven generation},
  author={Ruiz, Nataniel and Li, Yuanzhen and Jampani, Varun and Pritch, Yael and Rubinstein, Michael and Aberman, Kfir},
  booktitle={Proceedings of the IEEE/CVF conference on computer vision and pattern recognition},
  pages={22500--22510},
  year={2023}
}

@article{textual_inversion,
  title={An image is worth one word: Personalizing text-to-image generation using textual inversion},
  author={Gal, Rinon and Alaluf, Yuval and Atzmon, Yuval and Patashnik, Or and Bermano, Amit H and Chechik, Gal and Cohen-Or, Daniel},
  journal={arXiv preprint arXiv:2208.01618},
  year={2022}
}

@article{ip_adapter,
  title={Ip-adapter: Text compatible image prompt adapter for text-to-image diffusion models},
  author={Ye, Hu and Zhang, Jun and Liu, Sibo and Han, Xiao and Yang, Wei},
  journal={arXiv preprint arXiv:2308.06721},
  year={2023}
}

@article{ms_diffusion,
  title={MS-Diffusion: Multi-subject Zero-shot Image Personalization with Layout Guidance},
  author={Wang, X and Fu, Siming and Huang, Qihan and He, Wanggui and Jiang, Hao},
  journal={arXiv preprint arXiv:2406.07209},
  year={2024}
}

@inproceedings{emu2,
  title={Generative multimodal models are in-context learners},
  author={Sun, Quan and Cui, Yufeng and Zhang, Xiaosong and Zhang, Fan and Yu, Qiying and Wang, Yueze and Rao, Yongming and Liu, Jingjing and Huang, Tiejun and Wang, Xinlong},
  booktitle={Proceedings of the IEEE/CVF Conference on Computer Vision and Pattern Recognition},
  pages={14398--14409},
  year={2024}
}

@inproceedings{kosmosg,
title={Kosmos-G: Generating Images in Context with Multimodal Large Language Models},
author={Xichen Pan and Li Dong and Shaohan Huang and Zhiliang Peng and Wenhu Chen and Furu Wei},
booktitle={The Twelfth International Conference on Learning Representations},
year={2024},
url={https://openreview.net/forum?id=he6mX9LTyE}
}

@inproceedings{unsupervised_semantic_correspondence,
title={Unsupervised Semantic Correspondence Using Stable Diffusion},
author={Eric Hedlin and Gopal Sharma and Shweta Mahajan and Hossam Isack and Abhishek Kar and Andrea Tagliasacchi and Kwang Moo Yi},
booktitle={Thirty-seventh Conference on Neural Information Processing Systems},
year={2023},
url={https://openreview.net/forum?id=sovxUzPzLN}
}

@inproceedings{
emergent_correspondance,
title={Emergent Correspondence from Image Diffusion},
author={Luming Tang and Menglin Jia and Qianqian Wang and Cheng Perng Phoo and Bharath Hariharan},
booktitle={Thirty-seventh Conference on Neural Information Processing Systems},
year={2023},
url={https://openreview.net/forum?id=ypOiXjdfnU}
}

@misc{
generative_what,
title={Generative Models: What Do They Know? Do They Know Things? Let's Find Out!},
author={Xiaodan Du and Nicholas Kolkin and Greg Shakhnarovich and Anand Bhattad},
year={2025},
url={https://openreview.net/forum?id=xkR3bcswuC}
}

@inproceedings{editbench,
  title={Imagen editor and editbench: Advancing and evaluating text-guided image inpainting},
  author={Wang, Su and Saharia, Chitwan and Montgomery, Ceslee and Pont-Tuset, Jordi and Noy, Shai and Pellegrini, Stefano and Onoe, Yasumasa and Laszlo, Sarah and Fleet, David J and Soricut, Radu and others},
  booktitle={Proceedings of the IEEE/CVF conference on computer vision and pattern recognition},
  pages={18359--18369},
  year={2023}
}

@article{re_imagen,
  title={Re-imagen: Retrieval-augmented text-to-image generator},
  author={Chen, Wenhu and Hu, Hexiang and Saharia, Chitwan and Cohen, William W},
  journal={arXiv preprint arXiv:2209.14491},
  year={2022}
}

@article{knn_diffusion,
  title={Knn-diffusion: Image generation via large-scale retrieval},
  author={Sheynin, Shelly and Ashual, Oron and Polyak, Adam and Singer, Uriel and Gafni, Oran and Nachmani, Eliya and Taigman, Yaniv},
  journal={arXiv preprint arXiv:2204.02849},
  year={2022}
}

@article{retrieve_diffusion,
  title={Retrieval-augmented diffusion models},
  author={Blattmann, Andreas and Rombach, Robin and Oktay, Kaan and Muller, Jonas and Ommer, Bjorn},
  journal={Advances in Neural Information Processing Systems},
  volume={35},
  pages={15309--15324},
  year={2022}
}

@article{omnigen,
  title={Omnigen: Unified image generation},
  author={Xiao, Shitao and Wang, Yueze and Zhou, Junjie and Yuan, Huaying and Xing, Xingrun and Yan, Ruiran and Wang, Shuting and Huang, Tiejun and Liu, Zheng},
  journal={arXiv preprint arXiv:2409.11340},
  year={2024}
}

@article{RAG,
  title={Retrieval-augmented generation for knowledge-intensive nlp tasks},
  author={Lewis, Patrick and Perez, Ethan and Piktus, Aleksandra and Petroni, Fabio and Karpukhin, Vladimir and Goyal, Naman and Kuttler, Heinrich and Lewis, Mike and Yih, Wen-tau and Rocktaschel, Tim and others},
  journal={Advances in neural information processing systems},
  volume={33},
  pages={9459--9474},
  year={2020}
}

@inproceedings{vct,
  title={General image-to-image translation with one-shot image guidance},
  author={Cheng, Bin and Liu, Zuhao and Peng, Yunbo and Lin, Yue},
  booktitle={Proceedings of the IEEE/CVF international conference on computer vision},
  pages={22736--22746},
  year={2023}
}

@article{lambda_eclipse,
  title={lambda-ECLIPSE: Multi-Concept Personalized Text-to-Image Diffusion Models by Leveraging CLIP Latent Space},
  author={Patel, Maitreya and Jung, Sangmin and Baral, Chitta and Yang, Yezhou},
  journal={arXiv preprint arXiv:2402.05195},
  year={2024}
}

@inproceedings{tf_icon,
  title={Tf-icon: Diffusion-based training-free cross-domain image composition},
  author={Lu, Shilin and Liu, Yanzhu and Kong, Adams Wai-Kin},
  booktitle={Proceedings of the IEEE/CVF International Conference on Computer Vision},
  pages={2294--2305},
  year={2023}
}

@article{customcontrast,
  title={CustomContrast: A Multilevel Contrastive Perspective For Subject-Driven Text-to-Image Customization},
  author={Chen, Nan and Huang, Mengqi and Chen, Zhuowei and Zheng, Yang and Zhang, Lei and Mao, Zhendong},
  journal={arXiv preprint arXiv:2409.05606},
  year={2024}
}

@article{pixart_alpha,
  title={Pixart-alpha: Fast training of diffusion transformer for photorealistic text-to-image synthesis},
  author={Chen, Junsong and Yu, Jincheng and Ge, Chongjian and Yao, Lewei and Xie, Enze and Wu, Yue and Wang, Zhongdao and Kwok, James and Luo, Ping and Lu, Huchuan and others},
  journal={arXiv preprint arXiv:2310.00426},
  year={2023}
}

@inproceedings{pixart_sigma,
  title={Pixart-sigma: Weak-to-strong training of diffusion transformer for 4k text-to-image generation},
  author={Chen, Junsong and Ge, Chongjian and Xie, Enze and Wu, Yue and Yao, Lewei and Ren, Xiaozhe and Wang, Zhongdao and Luo, Ping and Lu, Huchuan and Li, Zhenguo},
  booktitle={European Conference on Computer Vision},
  pages={74--91},
  year={2024},
  organization={Springer}
}

@article{pixart_delta,
  title={Pixart-delta: Fast and controllable image generation with latent consistency models},
  author={Chen, Junsong and Wu, Yue and Luo, Simian and Xie, Enze and Paul, Sayak and Luo, Ping and Zhao, Hang and Li, Zhenguo},
  journal={arXiv preprint arXiv:2401.05252},
  year={2024}
}

@article{sdxl_lightning,
  title={Sdxl-lightning: Progressive adversarial diffusion distillation},
  author={Lin, Shanchuan and Wang, Anran and Yang, Xiao},
  journal={arXiv preprint arXiv:2402.13929},
  year={2024}
}

@article{ddim,
  title={Denoising diffusion implicit models},
  author={Song, Jiaming and Meng, Chenlin and Ermon, Stefano},
  journal={arXiv preprint arXiv:2010.02502},
  year={2020}
}

@article{dpm_sovler_plus,
  title={Dpm-solver++: Fast solver for guided sampling of diffusion probabilistic models},
  author={Lu, Cheng and Zhou, Yuhao and Bao, Fan and Chen, Jianfei and Li, Chongxuan and Zhu, Jun},
  journal={arXiv preprint arXiv:2211.01095},
  year={2022}
}

@article{rf_inversion,
  title={Semantic image inversion and editing using rectified stochastic differential equations},
  author={Rout, Litu and Chen, Yujia and Ruiz, Nataniel and Caramanis, Constantine and Shakkottai, Sanjay and Chu, Wen-Sheng},
  journal={arXiv preprint arXiv:2410.10792},
  year={2024}
}

@inproceedings{
lightningfast_inversion,
title={Lightning-Fast Image Inversion and Editing for Text-to-Image Diffusion Models},
author={Dvir Samuel and Barak Meiri and Haggai Maron and Yoad Tewel and Nir Darshan and Shai Avidan and Gal Chechik and Rami Ben-Ari},
booktitle={The Thirteenth International Conference on Learning Representations},
year={2025},
url={https://openreview.net/forum?id=t9l63huPRt}
}

@inproceedings{masactrl,
  title={Masactrl: Tuning-free mutual self-attention control for consistent image synthesis and editing},
  author={Cao, Mingdeng and Wang, Xintao and Qi, Zhongang and Shan, Ying and Qie, Xiaohu and Zheng, Yinqiang},
  booktitle={Proceedings of the IEEE/CVF international conference on computer vision},
  pages={22560--22570},
  year={2023}
}

@inproceedings{style_align,
  title={Style aligned image generation via shared attention},
  author={Hertz, Amir and Voynov, Andrey and Fruchter, Shlomi and Cohen-Or, Daniel},
  booktitle={Proceedings of the IEEE/CVF Conference on Computer Vision and Pattern Recognition},
  pages={4775--4785},
  year={2024}
}

@inproceedings{freeenhance,
  title={Freeenhance: Tuning-free image enhancement via content-consistent noising-and-denoising process},
  author={Luo, Yang and Zhang, Yiheng and Qiu, Zhaofan and Yao, Ting and Chen, Zhineng and Jiang, Yu-Gang and Mei, Tao},
  booktitle={Proceedings of the 32nd ACM International Conference on Multimedia},
  pages={7075--7084},
  year={2024}
}

@article{diffedit,
  title={Diffedit: Diffusion-based semantic image editing with mask guidance},
  author={Couairon, Guillaume and Verbeek, Jakob and Schwenk, Holger and Cord, Matthieu},
  journal={arXiv preprint arXiv:2210.11427},
  year={2022}
}

@article{attend_and_excite,
  title={Attend-and-excite: Attention-based semantic guidance for text-to-image diffusion models},
  author={Chefer, Hila and Alaluf, Yuval and Vinker, Yael and Wolf, Lior and Cohen-Or, Daniel},
  journal={ACM transactions on Graphics (TOG)},
  volume={42},
  number={4},
  pages={1--10},
  year={2023},
  publisher={ACM New York, NY, USA}
}

@inproceedings{ledits++,
  title={Ledits++: Limitless image editing using text-to-image models},
  author={Brack, Manuel and Friedrich, Felix and Kornmeier, Katharia and Tsaban, Linoy and Schramowski, Patrick and Kersting, Kristian and Passos, Apolin{\'a}rio},
  booktitle={Proceedings of the IEEE/CVF conference on computer vision and pattern recognition},
  pages={8861--8870},
  year={2024}
}

@inproceedings{blip2,
  title={BLIP-2: bootstrapping language-image pre-training with frozen image encoders and large language models},
  author={Li, Junnan and Li, Dongxu and Savarese, Silvio and Hoi, Steven},
  booktitle={Proceedings of the 40th International Conference on Machine Learning},
  pages={19730--19742},
  year={2023}
}

@ARTICLE{otsu,
  author={Otsu, Nobuyuki},
  journal={IEEE Transactions on Systems, Man, and Cybernetics}, 
  title={A Threshold Selection Method from Gray-Level Histograms}, 
  year={1979},
  volume={9},
  number={1},
  pages={62-66},
  doi={10.1109/TSMC.1979.4310076}}

@article{suti,
  title={Subject-driven text-to-image generation via apprenticeship learning},
  author={Chen, Wenhu and Hu, Hexiang and Li, Yandong and Ruiz, Nataniel and Jia, Xuhui and Chang, Ming-Wei and Cohen, William W},
  journal={Advances in Neural Information Processing Systems},
  volume={36},
  pages={30286--30305},
  year={2023}
}

@inproceedings{sdedit,
  title={SDEdit: Guided Image Synthesis and Editing with Stochastic Differential Equations},
  author={Meng, Chenlin and He, Yutong and Song, Yang and Song, Jiaming and Wu, Jiajun and Zhu, Jun-Yan and Ermon, Stefano},
  booktitle={International Conference on Learning Representations}
}

@article{p2p,
  title={Prompt-to-prompt image editing with cross attention control},
  author={Hertz, Amir and Mokady, Ron and Tenenbaum, Jay and Aberman, Kfir and Pritch, Yael and Cohen-Or, Daniel},
  journal={arXiv preprint arXiv:2208.01626},
  year={2022}
}

@inproceedings{pnp,
  title={Plug-and-play diffusion features for text-driven image-to-image translation},
  author={Tumanyan, Narek and Geyer, Michal and Bagon, Shai and Dekel, Tali},
  booktitle={Proceedings of the IEEE/CVF Conference on Computer Vision and Pattern Recognition},
  pages={1921--1930},
  year={2023}
}

@inproceedings{tis,
  title={Towards understanding cross and self-attention in stable diffusion for text-guided image editing},
  author={Liu, Bingyan and Wang, Chengyu and Cao, Tingfeng and Jia, Kui and Huang, Jun},
  booktitle={Proceedings of the IEEE/CVF conference on computer vision and pattern recognition},
  pages={7817--7826},
  year={2024}
}

@inproceedings{clip,
  title={Learning transferable visual models from natural language supervision},
  author={Radford, Alec and Kim, Jong Wook and Hallacy, Chris and Ramesh, Aditya and Goh, Gabriel and Agarwal, Sandhini and Sastry, Girish and Askell, Amanda and Mishkin, Pamela and Clark, Jack and others},
  booktitle={International conference on machine learning},
  pages={8748--8763},
  year={2021},
  organization={PmLR}
}

@article{imagereward,
  title={Imagereward: Learning and evaluating human preferences for text-to-image generation},
  author={Xu, Jiazheng and Liu, Xiao and Wu, Yuchen and Tong, Yuxuan and Li, Qinkai and Ding, Ming and Tang, Jie and Dong, Yuxiao},
  journal={Advances in Neural Information Processing Systems},
  volume={36},
  pages={15903--15935},
  year={2023}
}

@article{hpsv2,
  title={Human Preference Score v2: A Solid Benchmark for Evaluating Human Preferences of Text-to-Image Synthesis},
  author={Wu, Xiaoshi and Hao, Yiming and Sun, Keqiang and Chen, Yixiong and Zhu, Feng and Zhao, Rui and Li, Hongsheng},
  journal={CoRR},
  year={2023}
}

@article{dinov2,
  title={Dinov2: Learning robust visual features without supervision},
  author={Oquab, Maxime and Darcet, Timoth{\'e}e and Moutakanni, Th{\'e}o and Vo, Huy and Szafraniec, Marc and Khalidov, Vasil and Fernandez, Pierre and Haziza, Daniel and Massa, Francisco and El-Nouby, Alaaeldin and others},
  journal={arXiv preprint arXiv:2304.07193},
  year={2023}
}

@inproceedings{lpips,
  title={The unreasonable effectiveness of deep features as a perceptual metric},
  author={Zhang, Richard and Isola, Phillip and Efros, Alexei A and Shechtman, Eli and Wang, Oliver},
  booktitle={Proceedings of the IEEE conference on computer vision and pattern recognition},
  pages={586--595},
  year={2018}
}

@inproceedings{clip_dir,
  title={Diffusionclip: Text-guided diffusion models for robust image manipulation},
  author={Kim, Gwanghyun and Kwon, Taesung and Ye, Jong Chul},
  booktitle={Proceedings of the IEEE/CVF conference on computer vision and pattern recognition},
  pages={2426--2435},
  year={2022}
}
}


\end{document}